\documentclass[sigconf]{acmart}
\usepackage{caption}
\usepackage{graphicx}
\usepackage{subcaption}
\usepackage{hyperref}
\usepackage{amsfonts}
\usepackage{url}
\usepackage{multirow}
\usepackage{makecell}
\usepackage{color}
\usepackage{xspace}
\usepackage{amsmath}
\usepackage{adjustbox}
\usepackage{booktabs}
\usepackage{algorithm}
\usepackage{algpseudocode}
\usepackage{soul}
\usepackage{float}
\usepackage{color}
\usepackage{enumitem}

\newtheorem{theorem}{Theorem}[section]

\AtBeginDocument{%
  \providecommand\BibTeX{{%
    \normalfont B\kern-0.5em{\scshape i\kern-0.25em b}\kern-0.8em\TeX}}}

\copyrightyear{2023}
\acmYear{2023}
\setcopyright{acmlicensed}\acmConference[WWW '23]{Proceedings of the ACM Web Conference 2023}{May 1--5, 2023}{Austin, TX, USA}
\acmBooktitle{Proceedings of the ACM Web Conference 2023 (WWW '23), May 1--5, 2023, Austin, TX, USA}
\acmPrice{15.00}
\acmDOI{10.1145/3543507.3583544}
\acmISBN{978-1-4503-9416-1/23/04}

\begin{document}

%%
%% The "title" command has an optional parameter,
%% allowing the author to define a "short title" to be used in page headers.
\title{Toward Degree Bias in Embedding-Based \\ Knowledge Graph Completion}
\renewcommand{\shorttitle}{Toward Degree Bias in Embedding-Based KGC}

\author{Harry Shomer}
\email{shomerha@msu.edu}
\affiliation{%
  \institution{Michigan State University}
  \city{East Lansing}
  \country{USA}
}

\author{Wei Jin}
\email{jinwei2@msu.edu}
\affiliation{%
  \institution{Michigan State University}
  \city{East Lansing}
  \country{USA}
}

\author{Wentao Wang}
\email{wangw116@msu.edu}
\affiliation{%
  \institution{Michigan State University}
  \city{East Lansing}
  \country{USA}
}

\author{Jiliang Tang}
\email{tangjili@msu.edu}
\affiliation{%
  \institution{Michigan State University}
  \city{East Lansing}
  \country{USA}
}
% \author{Harry Shomer$^1$,~~~Wei Jin$^1$, ~~~Wentao Wang$^1$, and Jiliang Tang$^1$}
% \affiliation{
% 	\institution{
% 		$^1$Michigan State University, \{shomerha, jinwei2, wangw116, tangjili\}@msu.edu; \\
% 	}
% 	\country{}
% }
\renewcommand{\shortauthors}{Shomer, et al.}

\begin{abstract}
A fundamental task for knowledge graphs (KGs) is knowledge graph completion (KGC). It aims to predict unseen edges by learning representations for all the entities and relations in a KG. A common concern when learning representations on traditional graphs is degree bias. It can affect graph algorithms by learning poor representations for lower-degree nodes, often leading to low performance on such nodes. However, there has been limited research on whether there exists degree bias for embedding-based KGC and how such bias affects the performance of KGC. In this paper, we validate the existence of degree bias in embedding-based KGC and identify the key factor to degree bias. We then introduce a novel data augmentation method, KG-Mixup, to generate synthetic triples to mitigate such bias. Extensive experiments have demonstrated that our method can improve various embedding-based KGC methods and outperform other methods tackling the bias problem on multiple benchmark datasets. \footnote{The code is available at \url{https://github.com/HarryShomer/KG-Mixup}}
\end{abstract}

%%
%% The code below is generated by the tool at http://dl.acm.org/ccs.cfm.
%% Please copy and paste the code instead of the example below.
%%



\maketitle

\section{Introduction}

Knowledge graphs (KGs) are a specific type of graph where each edge represents a single fact. Each fact is represented as a triple $(h, r, t)$ that connects two entities $h$ and $t$ with a relation $r$. KGs have been widely used in many real-world applications such as  recommendation~\cite{cao2019unifying}, drug discovery~\cite{mohamed2020discovering}, and natural language understanding~\cite{liu2020k}. However, the  incomplete nature of KGs limits their applicability in those applications. To address this limitation, it is desired to perform a KG completion (KGC) task, i.e., predicting unseen edges in the graph thereby deducing new facts~\cite{rossi2021knowledge}. In recent years, the embedding-based methods~\cite{bordes2013translating,conve,tucker,rgcn} that embed a KG into a low-dimensional space have achieved remarkable success on KGC tasks and enable downstream applications.

\begin{figure}[t]
    \centering
    \includegraphics[width=0.5\columnwidth]{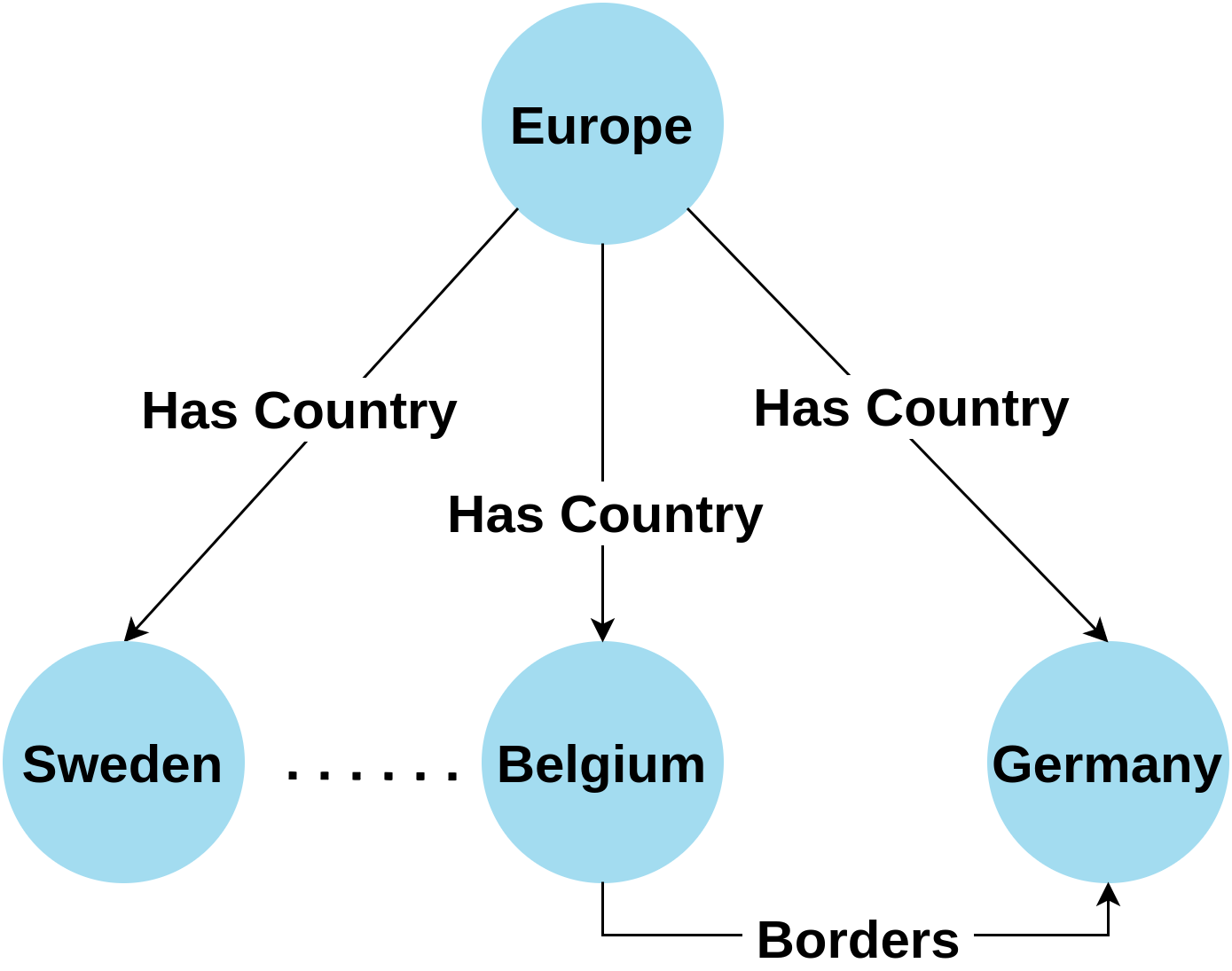}
    \caption{Example of multiple facts in a KG. Since each country only co-occurs with the relation \emph{Has Country} as the tail on one edge, they each only have a tail-relation degree of one with the relation  \emph{Has Country}. 
    }
    	\vspace{-0.15in}
    \label{fig:example}
\end{figure}

However, a common issue in graph-related tasks is degree bias~\cite{tang2020investigating,kojaku2021residual2vec}, where nodes of lower degree tend to learn poorer representations and have less satisfactory downstream performance.  Recent studies have validated this issue for various tasks  on homogeneous graphs such as  classification~\cite{tang2020investigating,graphsmote,liu2021tail} and link prediction~\cite{kojaku2021residual2vec}. However, KGs are naturally heterogeneous with multiple  types of nodes and relations. Furthermore, the study of degree bias on KGs is rather limited. Therefore, in this work, we ask \emph{whether degree bias exists in KGs and how it affects the model performance in the context of KGC}. 

To answer the aforementioned question, we perform preliminary studies to investigate how the degree affects the KGC performance. Take a triple $(h, r, t)$ as one example. The in-degree of entity $t$ is the number of triples where $t$ is the tail entity. Furthermore, we define the \emph{tail-relation degree} as the number of triples where $t$ and the relation $r$ co-occur as the tail and relation (Eq.~\eqref{eq:tail_rel_degree}). 
An example in Figure~\ref{fig:example} is the tail-relation pair (\emph{Germany}, \emph{Has Country}). Since the pair only co-occurs as a relation and tail in one triple, their tail-relation degree is 1.
Our preliminary studies (Section~\ref{sec:prelim}) suggest that when predicting the tail entity $t$, the in-degree of  $t$ and especially the tail-relation degree of $(t, r)$ plays a vital role. That is, when predicting the tail for a triple $(h, r, t)$, the number of triples where the entity $t$ and relation $r$ co-occur as an entity-relation pair correlates significantly with performance during KGC. Going back to our example, since \emph{Germany} and  \emph{Has Country} only co-occur as a relation and tail in one triple their tail-relation degree is low, thus making it difficult to predict \emph{Germany} for the query (\emph{Europe}, \emph{Has Country}, ?).

Given the existence of degree bias in KGC, we aim to alleviate the negative effect brought by degree bias. Specifically, we are tasked with improving the performance of triples with low tail-relation degrees while maintaining the performance of other triples with a higher tail-relation degree. Essentially, it is desired to promote the engagement of triples with low tail-relation degrees during training so as to learn better embeddings. To address this challenge, we propose a novel data augmentation framework. Our method works by augmenting entity-relation pairs that have low tail-relation degrees with synthetic triples. We generate the synthetic triples by extending the popular Mixup~\cite{mixup} strategy to KGs. Our contributions can be summarized as follows:
\begin{itemize}[leftmargin=0.1in]
    \item Through empirical study, we identify the degree bias problem in the context of KGC. To the best of our knowledge,\emph{ no previous work has studied the problem of degree bias from the perspective of entity-relation pairs}.
    \item We propose a simple yet effective data augmentation method, KG-Mixup, to alleviate the degree bias problem in KG embeddings.
    \item Through empirical analysis, we show that our proposed method can be formulated as a form of regularization on the low tail-relation degree samples.
    \item Extensive experiments have demonstrated that our proposed  method can improve the performance of lower tail-relation degree triples on multiple benchmark datasets without compromising the performance on triples of higher degree.
\end{itemize}

\section{Related Work}

\noindent \textbf{KG Embedding}: TransE~\cite{bordes2013translating} models the embeddings of a single triple as a translation in the embedding space. Multiple works model the triples as a tensor factorization problem, including~\cite{rescal, distmult, tucker, lowfer}. ConvE~\cite{conve} learns the embeddings by modeling the interaction of a single triple via a convolutional neural network. Other methods like R-GCN \cite{rgcn} modify GCN~\cite{kipf2017semi} for relational graphs. 
\vskip 0.5em
\noindent \textbf{Imbalanced/Long-Tail Learning}: Imbalanced/Long-Tail Learning considers the problem of model learning when the class distribution is highly uneven. SMOTE~\cite{chawla2002smote}, a classic technique, attempts to produce new synthetic samples for the minority class. Recent work has focused on tackling imbalance problems on deeper models. Works such as~\cite{ren2020balanced, tan2020equalization, lin2017focal} address this problem by modifying the loss for different samples. Another branch of work tries to tackle this issue by utilizing ensemble modeling~\cite{wang2020devil, zhou2020bbn, wang2020long}.  
\vskip 0.5em
\noindent \textbf{Degree Bias}: \citet{mohamed2020popularity} demonstrate the existence of popularity bias in popular KG datasets, which causes models to inflate the score of entities with a high degree. \citet{KG-biomedical} show the existence of entity degree bias in biomedical KGs.  \citet{rossi2021knowledge} demonstrate that the performance is positively  correlated with the number of source peers and negatively with the number of target peers. \citet{kojaku2021residual2vec} analyze the degree bias of random walks. To alleviate this issue, they propose a debiasing method that utilizes random graphs. In addition, many studies have focused on allaying the effect of degree bias for the task of node classification including \cite{tang2020investigating, graphsmote, liu2021tail}. 
However, there is no work that focuses on how the intersection of entity and relation degree bias effects embedding-based KGC.
\vskip 0.5em
\noindent \textbf{Data Augmentation for Graphs}
There is a line of works studying data augmentation  for homogeneous graphs~\cite{zhao2021data,zhao2022graph}. Few of these works study the link prediction problem~\cite{zhao2021counterfactual} but they do not address the issues in KGs.
To augment KGs, \citet{kg_data_aug} generate synthetic triples via adversarial learning; \citet{wang2018incorporating} use a GAN to create stronger negative samples; \citet{li2021rule} use rule mining to identify new triples to augment the graph with. 
However, all these methods do not augment with for the purpose of degree bias in KGs and hence are not applicable to the problem this paper studies.

\section{Preliminary Study} \label{sec:prelim}

\begin{figure*}[t]
\centering
    \begin{subfigure}{.33\textwidth}
      \centering
      \includegraphics[width=1.05\linewidth]{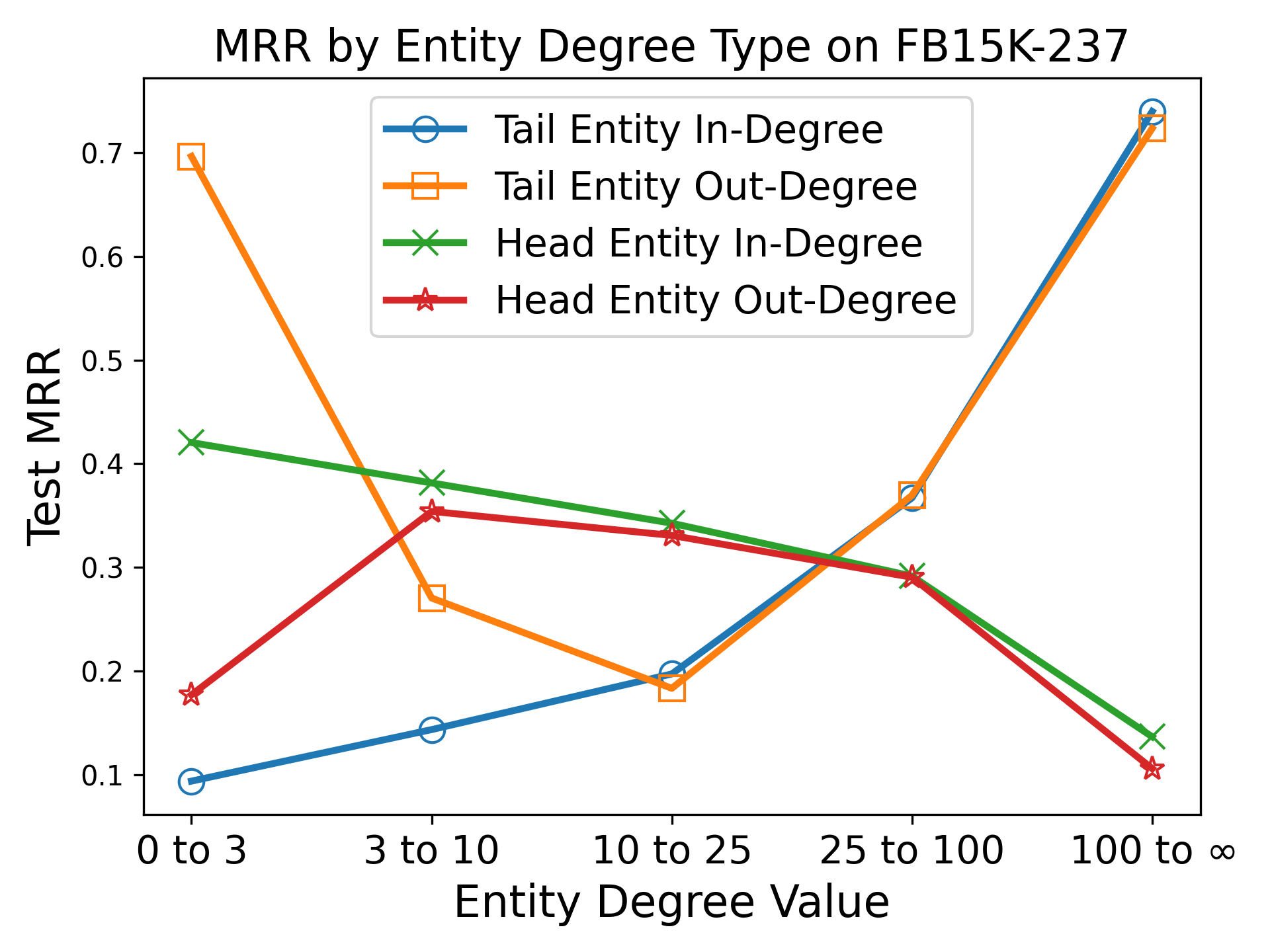}
      \caption{In and Out Degree Analysis}
      \vspace{-0.05in}
      \label{fig:tucker_ent_degree}
    \end{subfigure}%
    \begin{subfigure}{.33\textwidth}
      \centering
      \includegraphics[width=1.05\linewidth]{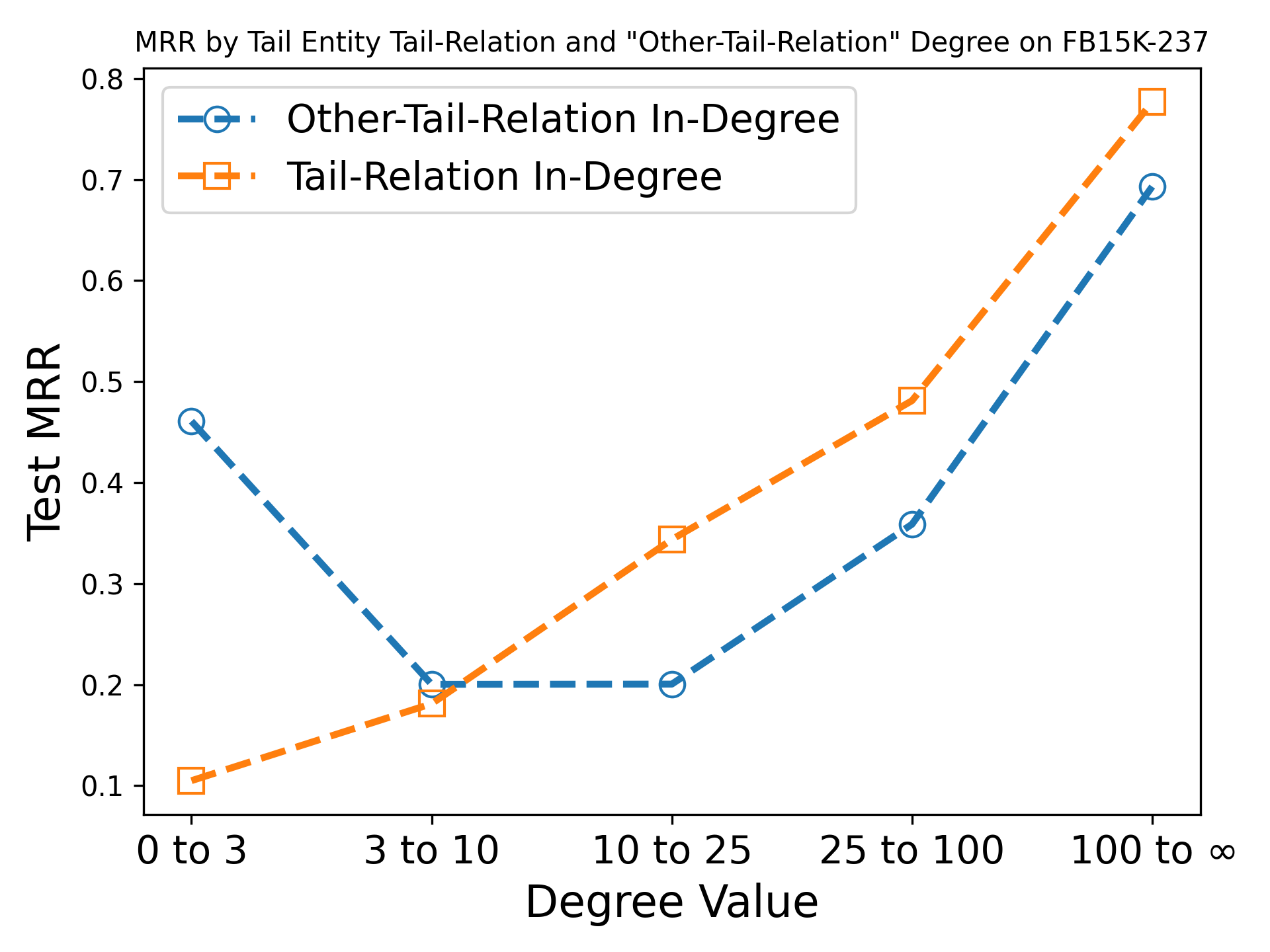}
      \caption{Tail In-Degree Analysis}
      \vspace{-0.05in}
      \label{fig:tucker_rel_specific_degree}
    \end{subfigure}%
    \begin{subfigure}{.33\textwidth}
      \centering
      \includegraphics[width=1.05\linewidth]{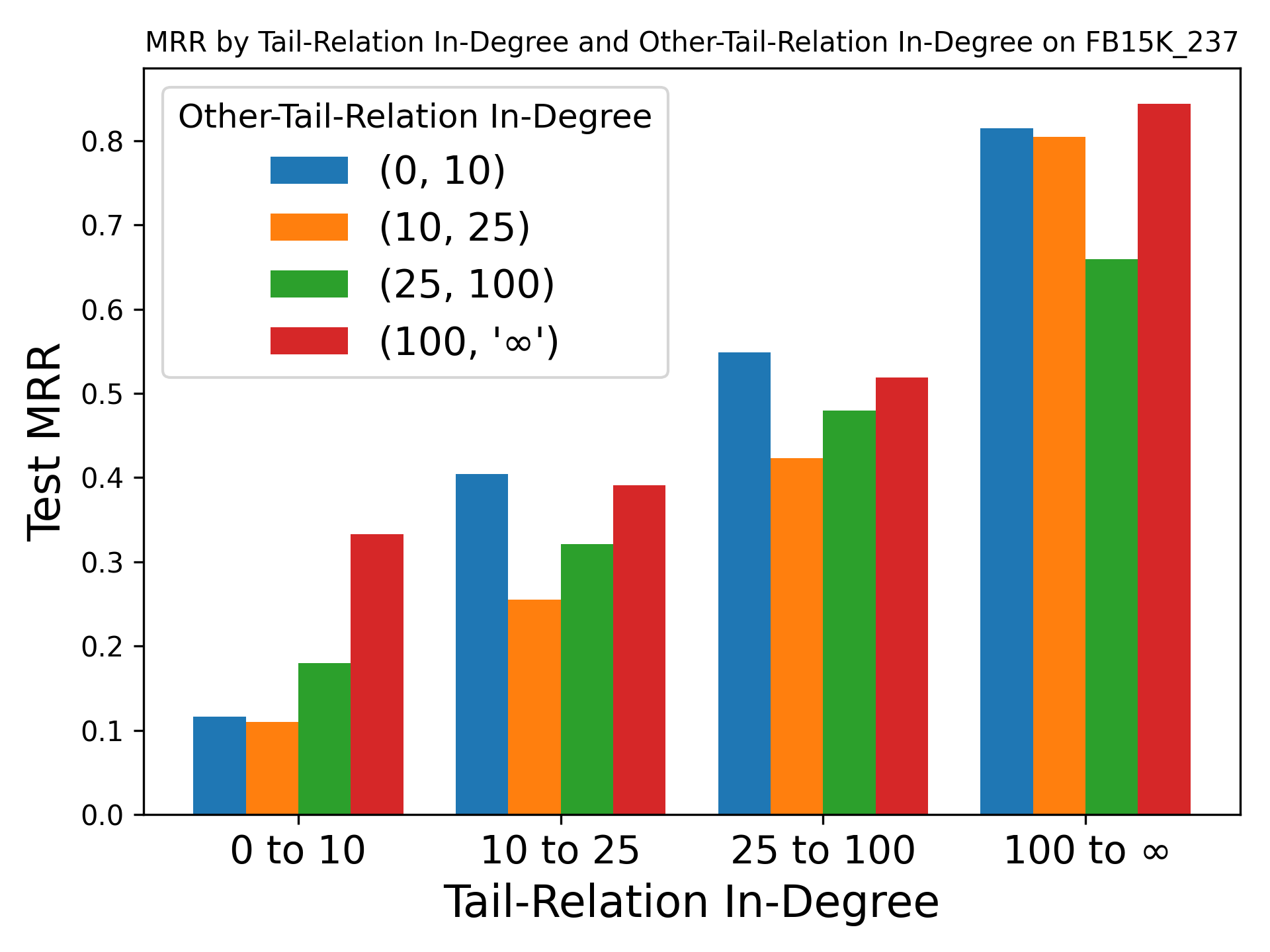}
      \caption{Tail-Relation vs. Other-Relation In-Degree}
      \vspace{-0.05in}
      \label{fig:tucker_control_degree}
    \end{subfigure}
    
    \captionsetup{width=1\linewidth}
    \caption{MRR when predicting the tail for TuckER on FB15K-237 when varying the (a) in-degree and out-degree of the head and tail entity, (b) tail-relation and other-relation in-degree, and (c) other-relation in-degree for smaller sub-ranges of the tail-relation degree. }
\vspace{-0.15in}

\label{fig:all_analysis_figs}
\end{figure*}

In this section we empirically study the degree bias problem in KGC. We focus on two representative embedding based methods ConvE~\cite{conve} and TuckER~\cite{tucker}. 
% {\it Note that there are recent path-based KGC methods~\cite{zhu2021neural} and we would like to leave the study of degree bias in these methods as one future work. }

We first introduce some notations. We denote $\mathcal{G}=\{\mathcal{V}, \mathcal{R}, \mathcal{E} \}$ as a KG with entities $\mathcal{V}$, relations $\mathcal{R}$,  and edges $\mathcal{E}$.
Each edge represents two entities connected by a single relation. We refer to an edge as a triple and denote it as $(h, r, t)$ where $h$ is referred to as the head entity, $t$ the tail entity, and $r$ the relation. Each entity and relation is represented by an embedding. We represent the embedding for a single entity $v$ as $\mathbf{x}_v \in \mathbb{R}^{n_v}$ and the embedding for a relation $r$ as $\mathbf{x}_r  \in \mathbb{R}^{n_r}$, where $n_v$ and $n_r$ are the dimensions of the entity and relation embeddings, respectively. We further define the degree of an entity $v$ as $d_v$ and the in-degree ($d_v^{(in)}$) and out-degree ($d_v^{(out)}$) as the number of triples where $v$ is the tail and head entity, respectively.

Lastly, KGC attempts to predict new facts that are not found in the original KG. This involves predicting the tail entities that satisfy $(h, r, ?)$ and the head entities that satisfy $(?, r, t)$. Following~\cite{dettmers2018convolutional}, we augment all triples $(h, r, t)$ with its inverse $(t, r^{-1}, h)$. As such, predicting the head entity of $(h, r, t)$ is analogous to predicting the tail entity for $(t, r^{-1}, ?)$. Under such a setting, KGC can be formulated as always predicting the tail entity. {\it Therefore, in the remainder of this work, we only consider KGC as predicting the tail entities that satisfy $(h, r, ?)$.}

In the following subsections, we will explore the following questions: (1) Does degree bias exist in typical KG embedding models? and (2) Which factor in a triple is related to such bias?  To answer these questions, we first study how the head and tail entity degree affect KGC performance in Section~\ref{sec:ent_analysis}. Then, we investigate the impact of the frequency of entity-relation pairs co-occurring on KGC performance in Section~\ref{sec:ent_rel_analysis}. 

\subsection{Entity Degree Analysis} \label{sec:ent_analysis}

We first examine the effect that the degree of both the head and tail entities have on KGC performance. We perform our analysis on the FB15K-237 dataset \cite{fb15k_237}, a  commonly used benchmark in KGC. Since a KG is a directed graph, we postulate that the direction of an entity's edges matters. We therefore split the degree of each entity into its in-degree and out-degree. We measure the performance using the mean reciprocal rank (MRR). Note that the degree metrics are calculated using only the training set.

Figure~\ref{fig:tucker_ent_degree} displays the results of TuckER (see Section~\ref{sec:train_procedure} for more details) on FB15K-237 split by both entities and degree type. 
From Figure~\ref{fig:tucker_ent_degree} we observe that when varying the tail entity degree value, the resulting change in test MRR is significantly larger than when varying the degree of head entities. Furthermore, the MRR increases drastically with the increase of tail in-degree (blue line) while there is a parabolic-like relationship when varying the tail out-degree (orange line). From these observations we can conclude: (1) the degree of the tail entity (i.e. the entity we are trying to predict) has a larger impact on test performance than the degree of the head entity;  (2) the tail in-degree features a more distinguishing and apparent relationship with performance than the tail out-degree.  
Due to the page limitation, the results of ConvE are shown in Appendix~\ref{sec:prelim_conve}, where we have similar observations.
These results suggest that KGC displays a degree bias in regards to the in-degree. Next, we will examine which factors of a triple majorly contribute to such degree bias.  

\subsection{Entity-Relation Degree Analysis} \label{sec:ent_rel_analysis}

In the previous subsection, we have demonstrated the relationship between the entity degree and KGC performance. %Next we aim to explore 
However, it doesn't account for the interaction of the entities and relation. Therefore, we further study how the presence of both the relation and entities in a triple \textit{together} impact the KGC performance. We begin by defining the number of edges that contains the relation $r$ and an entity $v$ as the \textit{relation-specific} degree: 
\begin{align} 
    d_{v, r} = \lvert \{(h, r, t) \in \mathcal{E} \: | \: h = v \lor t =v \} \rvert.
\end{align}
Based on the results in Section \ref{sec:ent_analysis}, we posit that the main indicator of performance is the in-degree of the tail entity. We extend this idea to our definition of relation-specific degree by only counting co-occurrences of an entity and relation when the entity occurs as the tail. For simplicity we refer to this as the \textit{tail-relation} degree and define it as:
\begin{equation} \label{eq:tail_rel_degree}
    d_{v, r}^{(tail)} = \lvert \{(h, r, v) \in \mathcal{E}\} \rvert.
\end{equation}
The tail-relation degree can be understood as the number of edges that an entity $v$ shares with $r$, where $v$ occupies the position we are trying to predict (i.e. the tail entity). We further refer to the number of in-edges that $v$ doesn't share with $r$ as ``Other-Tail Relation'' degree. This is calculated as the difference between the in-degree of entity $v$ and the tail-relation degree of $v$ and relation $r$, i.e. ${d_v^{(in)} - d_{v, r}^{(tail)}}$. It is easy to verify that the in-degree of an entity $v$ is the summation of the tail-relation degree and ``Other-Tail Relation'' degree.  We use Figure \ref{fig:example} as an example of the tail-relation degree. The entity \textit{Sweden} co-occurs with the relation \textit{Has Country} on one edge. On that edge, \textit{Sweden} is the tail entity. Therefore the tail-relation degree of the pair (\textit{Sweden}, \textit{Has Country}) is one. We note that a special case of the tail-relation degree is relation-level semantic evidence defined by~\citet{semantic_evidence}. 

Figure~\ref{fig:tucker_rel_specific_degree} displays the MRR when varying the value of the tail-relation and ``Other-Tail Relation'' degree of the tail entity. From the results, we note that while both degree metrics correlate with performance, the performance when the other-tail-relation degree in the range $[0, 3)$ is quite high. Since both metrics are highly correlated, it is difficult to determine which metric is more important for the downstream performance. Is the ``Other-Tail Relation'' the determining factor for performance or is it the tail-relation degree? We therefore check the performance when controlling for one another. Figure~\ref{fig:tucker_control_degree} displays the results when varying the ``Other-Tail Relation'' degree for specific sub-ranges of the tail-relation degree. From this figure, we see that the tail-relation degree exerts a much larger influence on the KGC performance as there is little variation between bars belonging to the same subset. Rather the tail-relation degree (i.e. the clusters of bars) has a much larger impact. Therefore, we conclude that for a single triple, the main factor of degree bias is the tail-relation degree of the tail entity.

\vskip 0.5em
\noindent\textbf{Remark.} Our analysis differs from traditional research on degree bias. While previous works focus only on the degree of the node, we focus on a specific type of frequency among entity-relation pairs. This is vital as the frequencies of both the entities and relations are important in KGs. Though we only analyze KGs, findings from our analysis could be applicable to other types of heterogeneous graphs.

\section{The Proposed Method} \label{sec:framework}

Grounded in the observations in Section~\ref{sec:ent_rel_analysis}, one natural idea to alleviate the degree bias in KGC is to compensate the triples with low tail-relation degrees. Based on this intuition, we propose a new method for  improving the KGC performance of triples with low tail-relation degrees. Our method, \textbf{KG-Mixup}, works by augmenting the low tail-relation degree triples during training with synthetic samples. This strategy has the effect of increasing the degree of an entity-relation pair with a low tail-relation degree by creating more shared edges between them. Therefore, KG-Mixup is very general and can further be used in conjunction with any KG embedding technique.

\subsection{General Problem}

In Section~\ref{sec:ent_rel_analysis} we showed that the tail-relation degree of the tail entity strongly correlates  with higher performance in KGC. As such we seek to design a method that can increase the performance of such low-degree entity-relation pairs without sacrificing the performance of high-degree pairs. 

To solve this problem, we consider data augmentation. Specifically, we seek to create synthetic triples for those entity-relations pairs with a low tail-relation degree. In such a way we are creating more training triples that contain those pairs, thereby ``increasing'' their degree. For each entity-relation pair with a tail-relation degree less than $\eta$, we add $k$ synthetic samples, which can be formulated  as follows:
\begin{equation} \label{eq:general_idea}
    \tilde{\mathcal{E}}_{v, r} = 
    \begin{cases} 
          \mathcal{E}_{v, r} \! \cup \! \{(\tilde{h}, \tilde{r}, \tilde{t})\}_{i=1}^k & d_{v, r}^{(tail)} < \eta, \\
          \mathcal{E}_{v, r} & \text{else}, \\
    \end{cases}
\end{equation}
where $(h, r, v) \in \mathcal{E}_{v, r}$ are the original training triples with the relation $r$ and the tail entity $v$, $(\tilde{h}, \tilde{r}, \tilde{t})$ is a synthetic sample, and $\tilde{\mathcal{E}}_{v, r}$ is the new set of triples to use during training.

%\subsection{Challenges}
\vskip 0.5em
\noindent\textbf{Challenges.} We note that creating the synthetic samples as shown in Eq.~\eqref{eq:general_idea} is non-trivial and there are a number of challenges: 
\begin{enumerate}[leftmargin=0.2in]
    \item How do we produce the synthetic samples for KG triples that contain multiple types of embeddings?
    \item How do we promote diversity in the synthetic samples $(\tilde{h}, \tilde{r}, \tilde{t})$? We want them to contain sufficient information from the original entity and relation embeddings we are augmenting, while also being distinct from similar triples in $\mathcal{E}_{v, r}$.
    \item How do we achieve such augmentation in a computationally efficient manner?
\end{enumerate}
These challenges motivate us to design a special data augmentation algorithm for knowledge graph completion and  we detail its core techniques in the next subsection.

\subsection{KG-Mixup}

\begin{algorithm}[t] \label{alg:kg_mixup}
\small
\captionsetup{font=small, labelfont=small}
\caption{KG-Mixup Training Procedure} \label{alg:kg_mixup}
\begin{algorithmic}[1]

\Require
    \Statex $G = \{V, R, \mathcal{E} \}$ \Comment{Training graph}
    \Statex $k, \eta$ \Comment{\# of samples to generate and degree threshold}
    \Statex $X, W$ \Comment{Model embeddings and parameters}

\State Randomly initialize $X$ and $W$
\State Pre-train to obtain initial $X$
\State Randomly re-initialize $W$

\While {not converged}
    \For{$e=(h_i, r_i, t_i) \in \mathcal{E}$}
        \If{$d_{t_i, r_i}^{(tail)} < \eta$} \label{lst:line:threshold}
            \State  $C = \{(h^*, r^*, t_i) \in \mathcal{E} \}$
            \State  $S = \text{Rand-Sample}(C, k)$
            \State  $\mathcal{E}_{\text{mix}} = \{\text{Mix}(e, s) \: \forall s \in S \}$ \Comment{Eq. \eqref{eq:kg_mixup}}
        \Else 
            \State $\mathcal{E}_{\text{mix}}= \{ \}$
        \EndIf
        \State Update model parameters on $\{e\} \cup \mathcal{E}_{\text{mix}}$
    \EndFor
\EndWhile
\State \Return $X$ and $W$
\end{algorithmic}
% \vspace{-0.1in}
\end{algorithm}

We now present our solution for producing synthetic samples as described in Eq.~\eqref{eq:general_idea}. Inspired by the popular Mixup~\cite{mixup} strategy,  we strive to augment the training set by mixing the representations of triples. We draw inspiration from mixup as (1) it is an intuitive and widely used data augmentation method, (2) it is able to promote diversity in the synthetic samples via the randomly drawn value $\lambda$, and (3) it is computationally efficient (see Section~\ref{sec:complexity}). 

We now briefly describe the general mixup algorithm. We denote the representations of two samples as $x_1$ and $x_2$ and their labels $y_1$ and $y_2$. Mixup creates a new sample $\tilde{x}$ and label $\tilde{y}$ by combining both the representations and labels via a random value $\lambda \in [0, 1]$ drawn from $\lambda \sim \text{Beta}(\alpha, \alpha)$ such that: 
\begin{align} \label{eq:original_mixup}
    &\tilde{x} = \lambda \mathbf{x}_1 + (1 - \lambda) \mathbf{x}_2,  \\
    &\tilde{y} = \lambda \mathbf{y}_1 + (1 - \lambda) \mathbf{y}_2. \label{eq:mix_y}
\end{align}
We adapt this strategy to our studied problem for a triple $(h, r, t)$ where the tail-relation degree is below a degree threshold, i.e. $d_{t, r}^{(tail)} < \eta$. For such a triple we aim to augment the training set by creating $k$ synthetic samples $\{(\tilde{h}, \tilde{r}, \tilde{t})\}_{i=1}^k$. This is done by mixing the original triple with $k$ other triples $\{(h_i, r_i, t_i)\}_{i=1}^k$.

However, directly adopting mixup to KGC leads to some problems: (1) Since each sample doesn't contain a label (Eq.~\ref{eq:mix_y}) we are unable to perform label mixing. (2) While standard mixup randomly selects samples to mix with, we may want to utilize a different selection criteria to better enhance those samples with a low tail-relation degree. (3) Since each sample is composed of multiple components (entities and relations) it's unclear how to mix two samples. We go over these challenges next. 

\subsubsection{Label Incorporation in KGC} We first tackle how to incorporate the label information as shown in~Eq.~\eqref{eq:mix_y}. Mixup was originally designed for classification problems, making the original label mixing straightforward. However, for KGC, we have no associated label for each triple. We therefore consider the entity we are predicting as the label. For a triple $e_1 = (h_1, r_1, t_1)$ where we are predicting $t_1$, the label would be considered the entity $t_1$. 

\subsubsection{Mixing Criteria} Per the original definition of Mixup, we would then mix $e_1$ with a triple belonging to the set $\{ (h_2, r_2, t_2) \in \mathcal{E} \: | \: t_2 \neq t_1\}$. However, since our goal is to predict $t_1$ we wish to avoid mixing it. Since we want to better predict $t_1$, we need to preserve as much tail (i.e. label) information as possible. As such, we only consider mixing with other triples that share the same tail and belong to the set $\{ (h_2, r_2, t_1) \in \mathcal{E}\ \: | \: h_1 \neq h_2, r_1 \neq r_2 \}$. Our design is similar to SMOTE \cite{chawla2002smote}, where only samples belonging to the same class are combined. We note that while it would be enticing to only consider mixing with triples containing the same entity-relation pairs, i.e. $(h_2, r_1, t_1) \in \mathcal{E}_{t_1, r_1}$, this would severely limit the number of possible candidate triples as the tail-relation degree can often be as low as one or two for some pairs.

\subsubsection{How to Mix?} We now discuss how to perform the mixing of two samples. Given a triple $e_1 = (h_1, r_1, t_1)$ of low tail-relation degree we mix it with another triple that shares the same tail  (i.e. label) such that $e_2 = (h_2, r_2, t_1)$.  Applying Eq. \eqref{eq:original_mixup} to $e_1$ and $e_2$, a synthetic triple $\tilde{e}=(\tilde{h}, \tilde{r}, \tilde{t})$ is equal to:
\begin{align} \label{eq:kg_mixup}
    &\tilde{e} = \lambda e_1 + (1-\lambda) e_2, \\
    &\tilde{e} = \lambda (h_1, r_1, t_1) + (1 - \lambda) (h_2, r_2, t_1), \\
    &\tilde{e} = \lambda (\mathbf{x}_{h_1}, \mathbf{x}_{r_1}, \mathbf{x}_{t_1}) + (1 - \lambda)  (\mathbf{x}_{h_2}, \mathbf{x}_{r_2}, \mathbf{x}_{t_1}),
\end{align}
where $\mathbf{x}_{h_i}$ and $\mathbf{x}_{r_j}$ represent the entity and relation embeddings, respectively. We apply the weighted sum to the head, relation, and tail, separately. Each entity and relation are therefore equal to:
\begin{align} \label{eq:kg_mixup_formula}
    &x_{\tilde{h}} = \lambda  \mathbf{x}_{h_1} + (1 - \lambda) \mathbf{x}_{h_2}, \\
    &x_{\tilde{r}} = \lambda  \mathbf{x}_{r_1} + (1 - \lambda) \mathbf{x}_{r_2}, \\
    &x_{\tilde{t}} = \mathbf{x}_{t_1}.
\end{align}
We use Figure \ref{fig:example} to illustrate an example. Let $e_1 = $ (\textit{Europe}, \textit{Has Country}, \textit{Germany}) be the triple we are augmenting. We mix it with another triple with the tail entity \textit{Germany}. We consider the triple $e_2 = $ (\textit{Belgium}, \textit{Borders}, \textit{Germany}). The mixed triple is represented as $\tilde{e} = (\textit{Europe} + \textit{Belgium}, \: \textit{Has Country} + \textit{Borders}, \: \textit{Germany})$. As $e_1$ contains the continent that $\textit{Germany}$ belongs to and $e_2$ has the country it borders, we can understand the synthetic triple $\tilde{e}$ as conveying the geographic location of \textit{Germany} inside of \textit{Europe}. This is helpful when predicting \textit{Germany} in the original triple $e_1$, since the synthetic sample imbues the representation of \textit{Germany} with more specific geographic information.

\subsection{KG-Mixup Algorithm for KGC}

We utilize the binary cross-entropy loss when training each model. The loss is optimized using the Adam optimizer~\cite{kingma2014adam}. We also include a hyperparameter $\beta$ for weighting the loss on the synthetic samples. The loss on a model with parameters $\theta$ is therefore:
\begin{equation}
    \mathcal{L}(\theta) = \mathcal{L}_{KG}(\theta) + \beta \mathcal{L}_{\text{Mix}}(\theta) \:,
\end{equation}
where $\mathcal{L}_{KG}$ is the loss on the original KG triples and $\mathcal{L}_{\text{Mix}}$ is the loss on the synthetic samples. The full algorithm is displayed in Algorithm \ref{alg:kg_mixup}. We note that we first pre-train the model before training with KG-Mixup, to obtain the initial entity and relation representations. This is done as it allows us to begin training with stronger entity and relation representations, thereby improving the generated synthetic samples.

\subsection{Algorithmic Complexity} \label{sec:complexity}

We denote the algorithmic complexity of a model $f$ (e.g. ConvE~\cite{conve} or TuckER~\cite{tucker}) for a single sample $e$ as $O(f)$. Assuming we generate $N$ negative samples per training sample, the training complexity of $f$ over a single epoch is:
\begin{equation}
    O \left( N \cdot \lvert \mathcal{E} \rvert \cdot O(f) \right),
\end{equation}
where $\lvert \mathcal{E} \rvert$ is the number of training samples. In KG-Mixup, in addition to scoring both the positive and negative samples, we also score the synthetic samples created for all samples with a tail-relation degree below a threshold $\eta$. We refer to that set of samples below the degree threshold as $\mathcal{E}_{\text{thresh}}$. We create $k$ synthetic samples per $e \in \mathcal{E}_{\text{thresh}}$. As such, our algorithm scores an additional $k \cdot \lvert \mathcal{E}_{\text{thresh}} \rvert$ samples for a total of $N \cdot \lvert \mathcal{E} \rvert + k \cdot \lvert \mathcal{E}_{\text{thresh}} \rvert$ samples per epoch. 
Typically the number of negative samples $N >> k$. Both ConvE and TuckER use all possible negative samples per training sample while we find $k=5$ works well. Furthermore, by definition, $\mathcal{E}_{\text{thresh}} \subseteq \mathcal{E}$ rendering $\lvert \mathcal{E} \rvert >> \lvert \mathcal{E}_{\text{thresh}} \rvert$. We can thus conclude that $N \cdot \lvert \mathcal{E} \rvert >> k \cdot \lvert \mathcal{E}_{\text{thresh}} \rvert$. We can therefore express the complexity of KG-Mixup as:
\begin{equation}
    % O(\text{KG-Mixup}) 
    \approx O \left( N \cdot \lvert \mathcal{E} \rvert \cdot O(f) \right).
\end{equation}
This highlights the efficiency of our algorithm as its complexity is approximately equivalent to the standard training procedure.

\section{Regularizing Effect of KG-Mixup} \label{sec:reg}

In this section, we examine the properties of KG-Mixup and show it can be formulated as a form of regularization on the entity and relation embeddings of low tail-relation degree samples following previous works \cite{carratino2020mixup, mixup_generalization}. 

We denote the mixup loss with model parameters $\theta$ over samples $S$ as $\mathcal{L}_{\text{Mix}}(\theta)$. The set $S$ contains those samples with a tail-relation degree below a threshold $\eta$ (see line~\ref{lst:line:threshold} in Algorithm \ref{alg:kg_mixup}). The embeddings for each sample $e_i = (h_i, r_i, t) \in S$ is mixed with those of a random sample $e_j = (h_j, r_j, t)$ that shares the same tail. The embeddings are combined via a random value $\lambda \sim \text{Beta}(\alpha, \alpha)$ as shown in Eq.~\eqref{eq:kg_mixup_formula}, thereby producing the synthetic sample $\tilde{e} = (\tilde{h}, \tilde{r}, t)$. The formulation for $\mathcal{L}_{\text{mix}}(\theta)$ is therefore:
\begin{equation} \label{eq:mixup_loss}
    \mathcal{L}_{\text{Mix}}(\theta) = \frac{1}{k \lvert S \rvert} \sum_{i=1}^{\lvert S \rvert} \sum_{j=1}^k l_{\theta} \left(\tilde{e}, \tilde{y} \right),
\end{equation}
where $k$ synthetic samples are produced for each sample in $S$, and $\tilde{y}$ is the mixed binary label. Following Theorem 1 in ~\citet{carratino2020mixup} we can rewrite the loss as the expectation over the synthetic samples as,
\begin{equation} \label{eq:mixup_carratino}
    \mathcal{L}_{\text{Mix}}(\theta) = \frac{1}{\lvert S \rvert} \sum_{i=1}^{\lvert S \rvert} \mathbb{E}_{\lambda, j} \: l_{\theta} \left(\tilde{e}, \tilde{y} \right),
\end{equation}
where $\lambda \sim \mathcal{D}_{\lambda}$ and $j \sim \text{Uniform}(\mathcal{E}_{t})$. The distribution $\mathcal{D}_{\lambda} = \text{Beta}_{[\frac{1}{2}, 1 ]}(\alpha, \alpha)$ and the set $\mathcal{E}_{t}$ contains all samples $(h_j, r_j, t)$ with tail $t$. Since the label $y$ for both samples $i$ and $j$ are always 1, rendering $\tilde{y}=1$, we can simplify Eq.~\eqref{eq:mixup_carratino} arriving at:
\begin{equation} \label{eq:mix_loss_last}
    \mathcal{L}_{\text{Mix}}(\theta) = \frac{1}{\lvert S \rvert} \sum_{i=1}^{\lvert S \rvert} \mathbb{E}_{\lambda, j} \: l_{\theta} \left( \tilde{e} \right).
\end{equation}
For the above loss function, we have the following theorem.
\begin{theorem} \label{th:mixup_reg_form}
The mixup loss $\mathcal{L}_{\text{Mix}}(\theta)$  defined in Eq.~\eqref{eq:mix_loss_last} can be rewritten as the following where the loss function $l_{\theta}$ is the binary cross-entropy loss,  $\mathcal{L} (\theta)$ is the loss on the original set of augmented samples $S$, and $\mathcal{R}_1 (\theta)$ and $\mathcal{R}_2 (\theta)$ are two regularization terms,
\begin{equation} \label{eq:mixup_lemma}
    \mathcal{L}_{\text{Mix}}(\theta) = \mathcal{L} (\theta) +  \mathcal{R}_1 (\theta) + \mathcal{R}_2 (\theta).
\end{equation}
The regularization terms are given by the following where each mixed sample $\tilde{e}$ is composed of a low tail-relation degree sample $e_i$ and another sample with the same tail entity $e_j$:  
\begin{align} \label{eq:reg_terms_theorem}
    &\mathcal{R}_1 (\theta) =  \frac{\tau}{\lvert S \rvert} \sum_{i=1}^{\lvert S \rvert} \sum_{j=1}^k  \left(1 - \sigma \left( f(e_i) \right)\right)  \frac{\partial f(e_i)^T}{\partial x_{\tilde{h}}} \Delta h, \\
    &\mathcal{R}_2 (\theta) = \frac{\tau}{\lvert S \rvert} \sum_{i=1}^{\lvert S \rvert} \sum_{j=1}^k  \left(1 - \sigma \left( f(e_i) \right)\right) \frac{\partial f(e_i)^T}{\partial x_{\tilde{r}}} \Delta r,
\end{align}
with $\tau = \mathbb{E}_{\lambda \sim \mathcal{D}_{\lambda}} (1-\lambda)$, $\Delta h = \left(x_{h_j} - x_{h_i}\right)$, $\Delta r = \left(x_{r_j} - x_{r_i}\right)$, $\sigma$ is the sigmoid function, and $f$ is the score function.
\end{theorem}

We provide the detailed proof of Theorem~\ref{th:mixup_reg_form} in Appendix~\ref{sec:appendix_proof}. Examining the terms in Eq~\eqref{eq:mixup_lemma},  we can draw the following understandings on KG-Mixup:
\begin{enumerate}[leftmargin=0.2in]
    \item The inclusion of $\mathcal{L} (\theta)$ implies that the low tail-relation degree samples are scored an additional time when being mixed. This can be considered as a form of oversampling on the low tail-relation degree samples.
    \item  If the probability is very high, i.e. $\sigma(f(e_i)) \approx 1$, both $\mathcal{R}_1$ and $\mathcal{R}_2$ cancel out. This is intuitive as if the current parameters perform well for the original low-degree sample, there is no need to make any adjustments. 
    \item \label{it:point2} 
    We can observe that $\mathcal{R}_1$ and $\mathcal{R}_2$ enforce some regularization on the derivatives as well as the difference between the embeddings $\Delta{h}$ and $\Delta{r}$. This motivates us to further examine the difference between the embeddings. In Section~\ref{sec:reg_exp}, we find that our method does indeed produce more similar embeddings, indicating that our method exerts a smoothing effect among mixed samples. 
    % \item \label{it:point2} In $\mathcal{R}_1$ and $\mathcal{R}_2$ there are two quantities minimized, (a) the gradient of the score function with respect to the mixed embeddings and (b) the difference between the embeddings of the two samples (i.e. $x_{h_j} - x_{h_i}$). The second minimized quantity, the difference in embeddings, prompts further analysis in Section~\ref{sec:reg_exp} showing that our method does indeed produce more similar embeddings. \harry{eh}
\end{enumerate}

\section{Experiment} \label{sec:experiments}

\begin{table*}[t] 
    \small
	\centering
	\caption{Knowledge Graph Completion (KGC) Comparison.}
\vskip -1em
%\vspace{-0.1in}
	\label{tab:main_results}
	\adjustbox{max width=\linewidth}{
		\begin{tabular}{@{}c | l | ccc | ccc | ccc @{}}
			\toprule
			 %  \multirow{1}{*}{\textbf{Method}} &
			 %  \multicolumn{1}{c}{\textbf{Method}} &
    		   \multicolumn{1}{c}{\textbf{Model}} &
    		   \multicolumn{1}{c}{\textbf{Method}} &
			   \multicolumn{3}{c}{\textbf{FB15K-237 }} & \multicolumn{3}{c}{\textbf{NELL-995}} & \multicolumn{3}{c}{\textbf{CoDEx-M}}  \\ 
			   \cmidrule(l){3-5} \cmidrule(l){6-8} \cmidrule(l){9-11} 
			   & & MRR   & H@1   & \multicolumn{1}{c}{H@10} & MRR   & H@1   & \multicolumn{1}{c}{H@10} & MRR   & H@1   & \multicolumn{1}{c}{H@10}  
			   \\ \midrule

            \multirow{5}{*}{\textbf{ConvE}} & Standard & \underline{33.04} & \underline{23.95} & \underline{51.23}    &  50.87 &	 \textbf{44.14} & 61.48    & {31.70}	& \textbf{24.34} & \underline{45.60} \\

            \cmidrule[0.5pt]{2-11}
            
            & + Over-Sampling & 30.45 & 21.85 & 47.81 & 48.63 & 40.99 & 60.78 & 27.13 & 20.17 & 40.11 \\
            & + Loss Re-weighting & 32.32 & 23.32 & 50.19 & \underline{50.89} & 43.83 & \underline{62.17} &  28.38 &  21.12 & 42.68 \\
            & + Focal Loss & 32.08 & 23.29 & 50.09 & 50.43 & \underline{44.00} & 60.70 & 27.99 & 20.93 & 41.48         \\
            & + KG-Mixup (Ours)  & \textbf{34.33} & \textbf{25.00} & \textbf{53.11} & \textbf{51.08} & 43.52 & \textbf{63.22} & \textbf{31.71} & \underline{23.49} & \textbf{47.49}          \\
            
            \midrule[1.25pt]
            
            \multirow{5}{*}{\textbf{TuckER}} & Standard &  35.19	& 26.06 & \underline{53.47} &    \underline{52.11} & 45.51 & \textbf{62.26} &     \underline{31.67} & \textbf{24.46} & \underline{45.73} \\

            \cmidrule[0.5pt]{2-11}
            
            & + Over-Sampling & 34.77 & 25.48 & 53.53 & 50.36 & 44.04 & 60.40 & 29.97 & 22.27 & 44.19 \\
            & + Loss Re-weighting & \underline{35.25} & \underline{26.08} & 53.34 & 51.91 & \underline{45.76} & 61.05 & 31.58 & \underline{24.32} & 45.41 \\
            & + Focal Loss & 34.02 & 24.79 & 52.48 &  49.57 & 43.28  & 58.91 & 31.47 & 24.05 & 45.60        \\
            & + KG-Mixup (Ours) & \textbf{35.83} & \textbf{26.37} & \textbf{54.78} & \textbf{52.24}  & \textbf{45.78} & \underline{62.14} & \textbf{31.90} & 24.15 & \textbf{46.54} \\
            
			\bottomrule
		\end{tabular}}
			%\vspace{-0.1in}
\end{table*}

\begin{table*}[t] 
    \small
	\centering
	\caption{MRR for tail-relation degree bins. The range for the zero, low, medium and high bins are [0, 1), [1, 10), [10, 50), and [50, $\infty$), respectively.}
     \vskip -1em
	%\vspace{-0.1in}
 	\label{tab:degree_results}
	\adjustbox{max width=\linewidth}{
            \begin{tabular}{@{}c | l | cccc | cccc | cccc @{}}			\toprule
			 %  \multirow{1}{*}{\textbf{Method}} &
    		   \multicolumn{1}{c}{\textbf{Model}} &
    		   \multicolumn{1}{c}{\textbf{Method}} &
			   \multicolumn{4}{c}{\textbf{FB15K-237 }} & \multicolumn{4}{c}{\textbf{NELL-995}} & \multicolumn{4}{c}{\textbf{CoDEx-M}}  \\ 
               \cmidrule(l){3-6} \cmidrule(l){7-10} \cmidrule(l){11-14}  
			   & & Zero & Low   & Medium   & \multicolumn{1}{c}{High} & Zero & Low   & Medium   & \multicolumn{1}{c}{High} & Zero & Low   & Medium   & \multicolumn{1}{c}{High}
			   \\ \midrule

            \multirow{5}{*}{\textbf{ConvE}} & Standard & 7.34 & 12.35 & \underline{34.95} & \textbf{70.97} & \underline{35.37} & 57.16 & \textbf{65.99} & \textbf{91.90} & 8.38 & \underline{7.97} & \textbf{34.64} & \textbf{65.29}  \\

            \cmidrule[0.5pt]{2-14}
            
            & + Over-Sampling      & 8.37 & \underline{12.45} & 33.01 & 68.75 & \textbf{36.67} & 57.33 & 56.09 & 79.57 & 8.09 & 7.52 & 29.51 & 54.80 \\
            & + Loss Re-weighting  & 5.03 & 9.89 & 30.56 & 63.34 & 36.16 & 57.96 & 63.69 & 89.52 & \underline{8.79} & 7.09 & 29.09  & 58.10 \\
            & + Focal Loss         & \underline{7.52} & 11.89 & 33.96 & 68.75 & 34.72 & \underline{58.00}  & \underline{65.60} & \underline{90.89} & 6.78 & 6.80 & \underline{33.42} & 56.96 \\
            & + KG-Mixup (Ours)    & \textbf{10.90} & \textbf{13.92} & \textbf{35.74} & \underline{70.72} & \underline{35.38} & \textbf{59.56}  & 65.41 & 90.64 & \textbf{9.74} & \textbf{8.96} & 32.63 & \underline{64.38} \\
            
            \midrule[1.25pt]
            
            \multirow{5}{*}{\textbf{TuckER}} & Standard & 10.41 & \underline{14.65} & \underline{38.49} & \underline{71.39} & \textbf{37.02} & 58.21 & \underline{69.17} & 90.55 & 9.99 & 8.29 & \textbf{35.23} & 63.94  \\
            
            \cmidrule[0.5pt]{2-14}
            
            & + Over-Sampling      & \textbf{12.25} & 14.28 & 36.79 & 70.50 & 34.50 & 55.46 & 65.68 & \textbf{93.47} & \textbf{10.98} & 7.76 &  32.50 & 60.25  \\
            & + Loss Re-weighting  & 10.61 & 14.40 & 37.66 & \textbf{72.28} & \underline{36.59} & \underline{59.00} &  67.19 & 91.17 &  \underline{10.44} & \underline{8.62} & \underline{35.00} & 63.39 \\
            & + Focal Loss         & 10.84 & 13.53 & 37.00 & 69.28 & 34.18 & 53.60 & 62.67 & 91.02 & 9.68 & 8.17 & 33.95 & \underline{64.13} \\
            & + KG-Mixup (Ours)    & \underline{11.83} & \textbf{15.61} & \textbf{39.45} & 70.86 & 36.12 & \textbf{60.73} & \textbf{71.67} & \underline{92.27} &  9.14 & \textbf{8.70} & 32.38 & \textbf{65.28}  \\
            
			\bottomrule
		\end{tabular}}
  \vspace{-0.1in}
\end{table*}

 In this section we conduct experiments to demonstrate the effectiveness of our approach on  multiple benchmark datasets. We further compare the results of our framework to other methods commonly used to address bias. In particular we study if KG-Mixup can (a) improve overall KGC performance and (b) increase performance on low tail-relation degree triples without degrading performance on other triples. We further conduct studies examining the effect of the regularization terms, ascertaining the importance of each component in our framework, and the ability of KG-Mixup to improve model calibration.

\subsection{Experimental Settings}

\subsubsection{Datasets} \label{sec:datasets}
We conduct experiments on three datasets including FB15K-237~\cite{fb15k_237}, CoDEx-M~\cite{codex}, and NELL-995~\cite{nell_995}. We omit the commonly used dataset WN18RR~\cite{conve} as a majority of entities have a degree less than or equal to 3, and as such does not exhibit any degree bias towards triples with a low tail-relation degree. The statistics of each dataset is shown in Table~\ref{tab:datasets}.

\subsubsection{Baselines}

We compare the results of our method, KG-Mixup, with multiple popular methods proposed for addressing imbalanced problems. Such methods can be used to mitigate bias caused by the initial imbalance. In our case, an imbalance in tail-relation degree causes algorithms to be biased against triples of low tail-relation degree. Specifically, we implement: (a) \textbf{Over-Sampling} triples below a degree threshold $\eta$. We over-sample $\eta - d_{v, r}^{(tail)}$ times, (b) \textbf{Loss Re-Weighting}~\cite{yuan2012sampling}, which assigns a higher loss to triples with a low tail-relation degree, (c) \textbf{Focal Loss}~\cite{lin2017focal}, which assigns a higher weight to misclassified samples (e.g. low degree triples).

\subsubsection{Evaluation Metrics}

To evaluate the model performance on the test set, we report the mean reciprocal rank (MRR) and the Hits@k for $k=1, 10$. Following~\cite{bordes2013translating}, we report the performance using the filtered setting. 

\subsubsection{Implementation Details} \label{sec:train_procedure}
In this section, we detail the training procedure used to train our framework KG-Mixup. We conduct experiments on our framework using two different KG embedding models, ConvE~\cite{conve} and TuckER~\cite{tucker}. Both methods are widely used to learn KG embeddings and serve as a strong indicator of our framework's efficacy. We use stochastic weight averaging (SWA)~\cite{swa} when training our model. SWA uses a weighted average of the parameters at different checkpoints during training for inference. Previous work~\cite{swa_aug} has shown that SWA in conjunction with data augmentation can increase performance. Lastly, the synthetic loss weighting parameter $\beta$ is determined via hyperparameter tuning on the validation set.

\subsection{Main Results}

In this subsection we evaluate KG-Mixup on multiple benchmarks, comparing its test performance against the baseline methods. We first report the overall performance of each method on the three datasets. We then report the performance for various degree bins. The top results are bolded with the second best underlined. Note that the \textbf{Standard} method refers to training without any additional method to alleviate bias.

Table \ref{tab:main_results} contains the overall results on each method and dataset. The performance is reported for both ConvE and TuckER. KG-Mixup achieves for the best MRR and Hits@10 on each dataset for ConvE. For TuckER, KG-Mixup further achieves the best MRR on each dataset and the top Hits@10 for two. Note that the other three baseline methods used for alleviating bias, on average, perform poorly. This may be due to their incompatibility with relational structured data where each sample contains multiple components. It suggests that we need dedicated efforts to handle the degree bias in KGC. 

We further report the MRR of each method for triples of different tail-relation degree. We split the triples into four degree bins of zero, low, medium and high degree. The range of each bin is [0, 1), [1, 10], [10, 50), and [50, $\infty$), respectively. KG-Mixup achieves a notable increase in performance on low tail-relation degree triples for each dataset and embedding model. KG-Mixup increases the MRR on low degree triples by 9.8\% and 5.3\%  for ConvE and TuckER, respectively, over the standard trained models on the three datasets. In addition to the strong increase in low degree performance, KG-Mixup is also able to retain its performance for high degree triples. The MRR on high tail-relation degree triples degrades, on average, only 1\% on ConvE between our method and standard training and actually increases 1\% for TuckER. Interestingly, the performance of KG-Mixup on the triples with zero tail-relation degree isn't as strong as the low degree triples. We argue that such triples are more akin to the zero-shot learning setting and therefore different from the problem we are studying. 

Lastly, we further analyzed the improvement of KG-Mixup over standard training by comparing the difference in performance between the two groups via the paired t-test. We found that for the results in Table~\ref{tab:main_results}, $5/6$ are statistically significant (p<0.05). Furthermore, for the performance on low tail-relation degree triples in Table~\ref{tab:degree_results}, all results ($6/6$) are statistically significant. This gives further justification that our method can improve both overall and low tail-relation degree performance.

\subsection{Regularization Analysis} \label{sec:reg_exp}

In this subsection we empirically investigate the regularization effects of KG-Mixup discussed in Section~\ref{sec:reg}. In Section~\ref{sec:reg} we demonstrated that KG-Mixup can be formulated as a form of regularization. 
We further showed that one of the quantities minimized is the difference between the head and relation embeddings of the two samples being mixed, $e_i$ and $e_j$, such that $(x_{h_j} - x_{h_i})$ and $(x_{r_j} - x_{r_i})$. Here $e_i$ is the low tail-relation degree sample being augmented and $e_j$ is another sample that shares the same tail. We deduce from this that for low tail-relation degree samples, KG-Mixup may cause their head and relation embeddings to be more similar to those of other samples that share same tail. Such a property forms a smoothing effect on the mixed samples, which facilitates a transfer of information to the embeddings of the low tail-relation degree sample.

%which is helpful  for the low-degree triplets as their information can be noisy.
% Such a property is desirable as the embeddings may become more similar, thereby facilitating a transfer of information to the embeddings of the low tail-relation degree sample. \harry{Is this ok? More? Less? idc}

We investigate this by comparing the head and relation embeddings of all samples that are augmented with all the head and relation embeddings that also share the same tail entity.
We denote the set of all samples below some tail-relation degree threshold $\eta$ as $\mathcal{E}_{\text{thresh}}$  and all samples with tail entity $t$ as $\mathcal{E}_t$. Furthermore, we refer to all head entities that are connected to a tail $t$ as $\mathcal{H}_t = \{h_j \; | \; (h_j, r_j, t) \in \mathcal{E}_t\}$ and all such relations as $\mathcal{R}_t = \{r_j \: | \: (h_j, r_j, t) \in \mathcal{E}_t\}$. For each sample $(h_i, r_i, t) \in \mathcal{E}_{\text{thresh}}$ we compute the mean euclidean distance between the (1) head embedding $\mathbf{x}_{h_i}$ and all $\mathbf{x}_{h_j} \in \mathcal{H}_t$ and (2) the relation embedding $\mathbf{x}_{r_i}$ and all $\mathbf{x}_{r_j} \in \mathcal{R}_t$. For a single sample $e_i$ the mean head and relation embedding distance are given by $h_{\text{dist}}(e_i)$ and $r_{\text{dist}}(e_i)$, respectively. Lastly, we take the mean of both the head and relation embeddings mean distances across all $e \in \mathcal{E}_{\text{thresh}}$, 
\begin{align}
    &D_{\text{rel}} = \text{Mean} \left(r_{\text{dist}}(e_i) \: | \: e_i \in \mathcal{E}_{\text{thresh}} \right), \\
    &D_{\text{head}} = \text{Mean} \left(h_{\text{dist}}(e_i) \: | \: e_i \in \mathcal{E}_{\text{thresh}} \right).
\end{align}
Both $D_{\text{head}}$ and $D_{\text{rel}}$ are shown in Table~\ref{tab:emb_dist} for models fitted with and without KG-Mixup. We display the results for ConvE on FB15K-237. For both the mean head and relation distances, KG-Mixup produces smaller distances than the standardly-trained model. This aligns with our previous theoretical understanding of the regularization effect of the proposed method: for samples for which we augment during training, their head and relation embeddings are more similar to those embeddings belonging to other samples that share the same tail. This to some extent forms a smoothing effect, which is helpful for learning better representations for the low-degree triplets.

\begin{table}[h]
\small
\centering
    \caption{Mean Embedding Distances on FB15K-237.}
    \vskip -1em
    \begin{tabular}{@{}l | c | c @{}}
        \toprule
            \textbf{Embedding Type} &
            \textbf{Head Entity} &
            \textbf{Relation}  \\ 
        \midrule
        w/o KG-Mixup & 1.18 & 1.21 \\
        KG-Mixup     & 1.09 & 1.13 \\
        \midrule[0.75pt]
        \% Decrease  & -7.6\% & -6.6\% \\
        \bottomrule
    \end{tabular}
\label{tab:emb_dist}
    \vskip -1em
\end{table}

\subsection{Ablation Study}

In this subsection we conduct an ablation study of our method on the FB15K-237 dataset using ConvE and TuckER. We ablate both the data augmentation strategy and the use of stochastic weight averaging (SWA) separately to ascertain their effect on performance. We report the overall test MRR and the low tail-relation degree MRR. The results of the study are shown in Table~\ref{tab:ablation}. KG-Mixup achieves the best overall performance on both embedding models. Using only our data augmentation strategy leads to an increase in both the low degree and overall performance. On the other hand, while the SWA-only model leads to an increase in overall performance it degrades the low degree performance. We conclude from these observations that data augmentation component of KG-Mixup is vital for improving low degree performance while SWA helps better maintain or even improve performance on the non-low degree triples.

\begin{table}[h]
    \small
	\centering
	\caption{Ablation Study on FB15K-237.}
     \vskip -1em
	\label{tab:ablation}
	\adjustbox{max width=\linewidth}{
		\begin{tabular}{@{}l | cc | cc @{}}
			\toprule
    		   \multicolumn{1}{c}{\textbf{Method}} &
			   \multicolumn{2}{c}{\textbf{ConvE}} &
			   \multicolumn{2}{c}{\textbf{TuckER}} \\ 
			    \cmidrule(l){2-3} \cmidrule(l){4-5} 
			   & Low  & Overall & Low & Overall 
			   \\ \midrule
               Standard & 12.35 & 33.04 & 14.65 & 35.19 \\
               + SWA   & 12.27 & 33.69 &  14.18 &  35.77 \\
               + Augmentation &  \textbf{13.99} &  33.67 &  \textbf{15.64} & 35.62  \\
               KG-Mixup (Ours) & 13.92 & \textbf{34.33} & 15.61 & \textbf{35.83}  \\
			\bottomrule
		\end{tabular}}
  \vspace{-0.1in}
\end{table}

\subsection{Parameter Study}

\begin{figure}[t]
\centering
    \begin{subfigure}{.25\textwidth}
      \centering
      \includegraphics[width=0.95
      \linewidth]{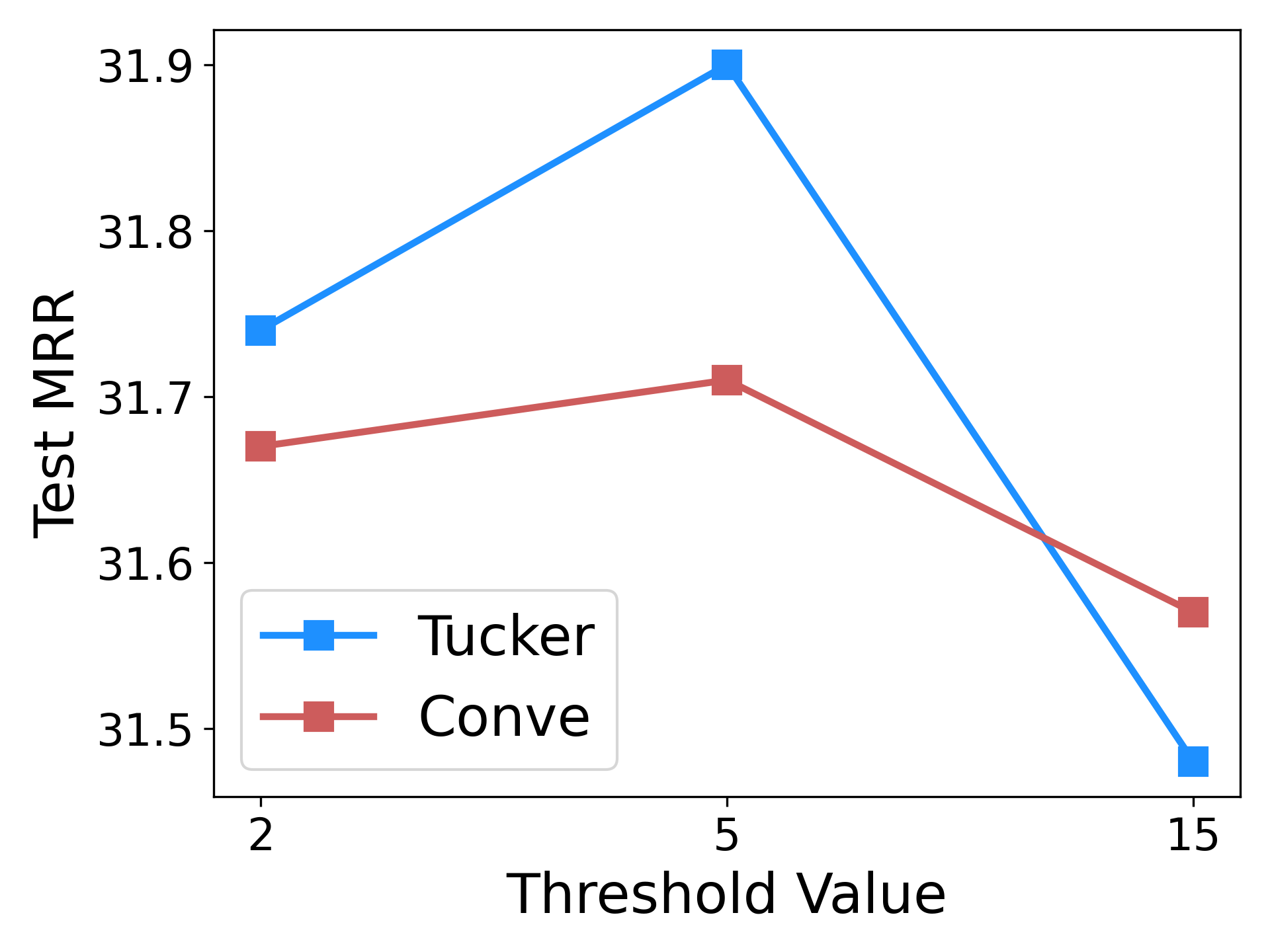}
      \caption{Varying degree threshold}
      \label{fig:thresh_study}
    \end{subfigure}%
    \begin{subfigure}{.25\textwidth}
      \centering
    \includegraphics[width=0.95\linewidth]{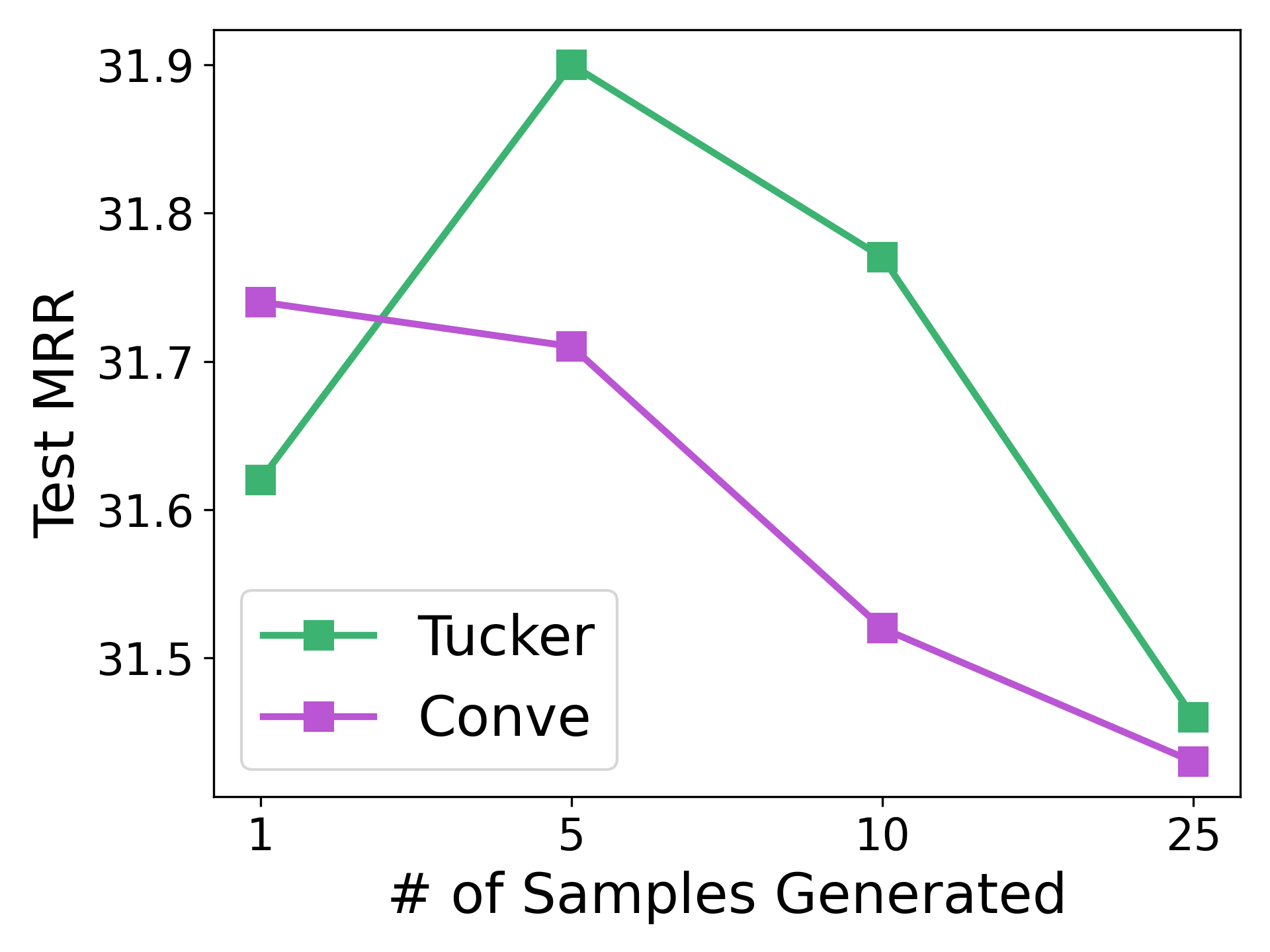}
      \caption{Varying  \# of samples}
      \label{fig:gen_study}
    \end{subfigure}%
    	%\vspace{-0.1in}
\caption{MRR of TuckER and ConvE on CoDEx-M (a) when varying the degree threshold and (b) when varying the number of samples generated.}
\label{fig:param_study}
	\vspace{-0.1in}
\end{figure}

In this subsection we study how varying the number of generated synthetic samples $k$ and the degree threshold $\eta$ affect the performance of KG-Mixup. We consider the values  $k \in \{1, 5, 10, 25 \}$ and $\eta \in \{2, 5, 15 \}$. We report the MRR for both TuckER and ConvE on the CoDEx-M dataset. Figure  \ref{fig:thresh_study} displays the performance when varying the degree threshold. Both methods peak at a value of $\eta=5$ and perform worst at $\eta=15$. Figure \ref{fig:gen_study} reports the MRR when varying the number of synthetic samples generated. Both methods peak early with ConvE actually performing best at $k=1$. Furthermore, generating too many samples harms performance as evidenced by the sharp drop in MRR occurring after $k=5$.

\subsection{Model Calibration}
In this subsection we demonstrate that KG-Mixup is effective at improving the calibration of KG embedding models. Model calibration~\cite{guo2017calibration} is concerned with how well calibrated a models prediction probabilities are with its accuracy. 
% That is, we expect that the produced probabilities be suggestive of the actual performance. 
Previous work~\cite{pleiss2017fairness} have discussed the desirability of calibration to minimize bias between different groups in the data (e.g. samples of differing degree). Other work~\cite{wald2021calibration} has drawn the connection between out-of-distribution generalization and model calibration, which while not directly applicable to our problem is still desirable. Relevant to our problem,~\citet{thulasidasan2019mixup} has shown that Mixup is effective at calibrating deep models for the tasks of image and text classification. As such, we investigate if KG-Mixup is helpful at calibrating KG embedding models for KGC. 

We compared the expected calibration error (see Appendix~\ref{sec:ece} for more details) between models trained with KG-Mixup and those without on multiple datasets. We report the calibration in Table~\ref{tab:calibration_results} for all samples and those with a low tail-relation degree. We find that in every instance KG-Mixup produces a better calibrated model for both ConvE and TuckER. These results suggest another reason for why KG-Mixup works; a well-calibrated model better minimizes the bias between different groups in the data~\cite{pleiss2017fairness}. This is integral for our problem where certain groups of data (i.e. triples with low tail-relation degree) feature bias. 

\begin{table}[h] 
    % \small
	\centering
	\caption{Expected Calibration Error (ECE). Lower is better.}
     \vskip -1em
	\label{tab:calibration_results}
	\adjustbox{max width=\linewidth}{
		\begin{tabular}{@{}l | c | cc | cc | cc @{}}
			\toprule
    		   \multicolumn{1}{c}{\textbf{Model}} &
    		   \multicolumn{1}{c}{\textbf{Method}} &
			   \multicolumn{2}{c}{\textbf{FB15K-237 }} & \multicolumn{2}{c}{\textbf{NELL-995}} & \multicolumn{2}{c}{\textbf{CoDEx-M}}  \\ 
			   \cmidrule(l){3-4} \cmidrule(l){5-6} \cmidrule(l){7-8} 
			   & & Low & Overall    & Low & Overall  & Low & Overall
			   \\ \midrule

            \multirow{2}{*}{\textbf{ConvE}} & Standard &  0.19 & 0.15 & 0.34 & 0.27 & 0.28 & 0.26 \\

            & KG-Mixup & 0.08 & 0.05 & 0.08 & 0.08 & 0.02 & 0.09 \\
            
            \midrule[1pt]
            
            \multirow{2}{*}{\textbf{TuckER}} & Standard & 0.20 & 0.35 & 0.63 & 0.56 & 0.05 & 0.34 \\
            
            & KG-Mixup &  0.07 & 0.1 & 0.26 & 0.20 & 0.01 & 0.06\\
            
			\bottomrule
		\end{tabular}}
		\vspace{-0.1in}
\end{table}
\section{Conclusion}

We explore the problem of degree bias in KG embeddings. Through empirical analysis we find that when predicting the tail $t$ for a triple $(h, r, t)$, a strong indicator performance is the number of edges where $r$ and $t$ co-occur as the relation and tail, respectively. We refer to this as the tail-relation degree. We therefore propose a new method, KG-Mixup, that can be used in conjunction with any KG embedding technique to improve performance on triples with a low tail-relation degree. It works by augmenting lower degree entity-relation pairs with additional synthetic triples during training. To create synthetic samples we adapt the Mixup~\cite{mixup} strategy to KGs. Experiments validate its usefulness. For future work we plan on expanding our method to path-based techniques such as NBFNet~\cite{zhu2021neural}. 

\begin{acks}
This research is supported by the National Science Foundation (NSF) under grant numbers CNS1815636, IIS1845081, IIS1928278, IIS1955285, IIS2212032, IIS2212144, IOS2107215, and IOS2035472, the Army Research Office (ARO) under grant number W911NF-21-1-0198, the Home Depot, Cisco Systems Inc, Amazon Faculty Award, Johnson\&Johnson and SNAP.
\end{acks}

\bibliography{ref}
\bibliographystyle{ACM-Reference-Format}

% \newpage
\appendix
\appendix
\section{Appendix} \label{sec:appendix}

\subsection{Dataset Statistics}

The statistics for each dataset can be found in Table~\ref{tab:datasets}.

\begin{table}[h]
\small
\centering
    \caption{Dataset Statistics.}
    \begin{tabular}{@{}lccc@{}}
        \toprule
            \textbf{Statistic} &
            \textbf{FB15K-237} & 
            \textbf{NELL-995}  & 
            \textbf{CoDEx-M}   \\ 
        \midrule
        \#Entities   & 14,541 & 74,536 & 17,050 \\
        \#Relations  & 237 & 200 & 51 \\
        \#Train      & 272,115 & 149,678 & 185,584 \\
        \#Validation & 17,535 & 543 & 10,310 \\
        \#Test       & 20,466 & 2,818 & 10,311 \\
        \bottomrule
    \end{tabular}
\label{tab:datasets}
\end{table}

\subsection{Infrastructure}

All experiments were run on a single 32G Tesla V100 GPU and implemented using PyTorch \cite{NEURIPS2019_9015}.

\subsection{Parameter Settings}
Each model is trained for 400 epochs on FB15K-237, 250 epochs on CoDEx-M, and 300 epochs on NELL-995. The embedding dimension is set to 200 for both methods except for the dimension of the relation embeddings in TuckER which is is tuned from $\{50, 100, 200 \}$. The batch size is set to 128 and the number of negative samples per positive sample is 100. The learning rate is tuned from $\{1e^{-5}, 5e^{-5}, 1e^{-4}, 5e^{-4}, 1e^{-3}\}$, the decay from $\{0.99, 0.995, 1 \}$, the label smoothing from $\{0, 0.1 \}$, and the dropout from \{0, 0.1, 0.2, 0.3, 0.4, 0.5 \}. For KG-Mixup we further tune  the degree threshold from $\{3, 5, 10 \}$, the number of samples generated from $\{5, 10 \}$, and the loss weight for the synthetic samples from \{1e-2, 1e-1, 1 \}. Lastly, we tune the stochastic weight averaging (SWA) initial learning rate from \{ 1e-5, 5e-4 \}. The best hyperparameter values for ConvE and TuckER using KG-Mixup are shown in Figures~\ref{tab:conve_hyperparam} and~\ref{tab:tucker_hyperparam}, respectively.

\subsection{Preliminary Study Results using ConvE} \label{sec:prelim_conve}

\begin{figure*}[t]
\centering
    \begin{subfigure}{.33\textwidth}
      \centering
      \includegraphics[width=1\linewidth]{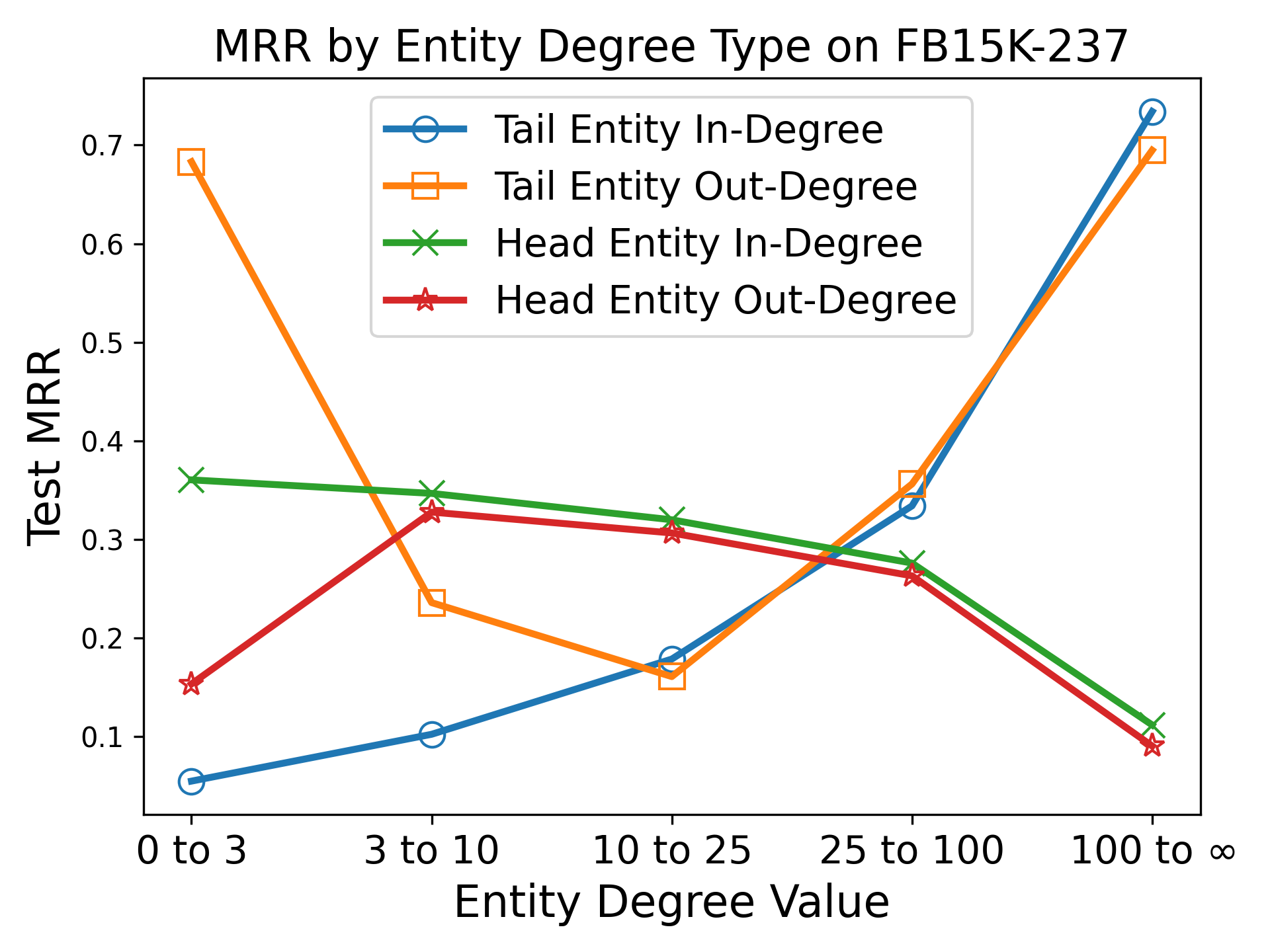}
      \caption{}
      \label{fig:conve_ent_degree}
    \end{subfigure}%
    \begin{subfigure}{.33\textwidth}
      \centering
      \includegraphics[width=1\linewidth]{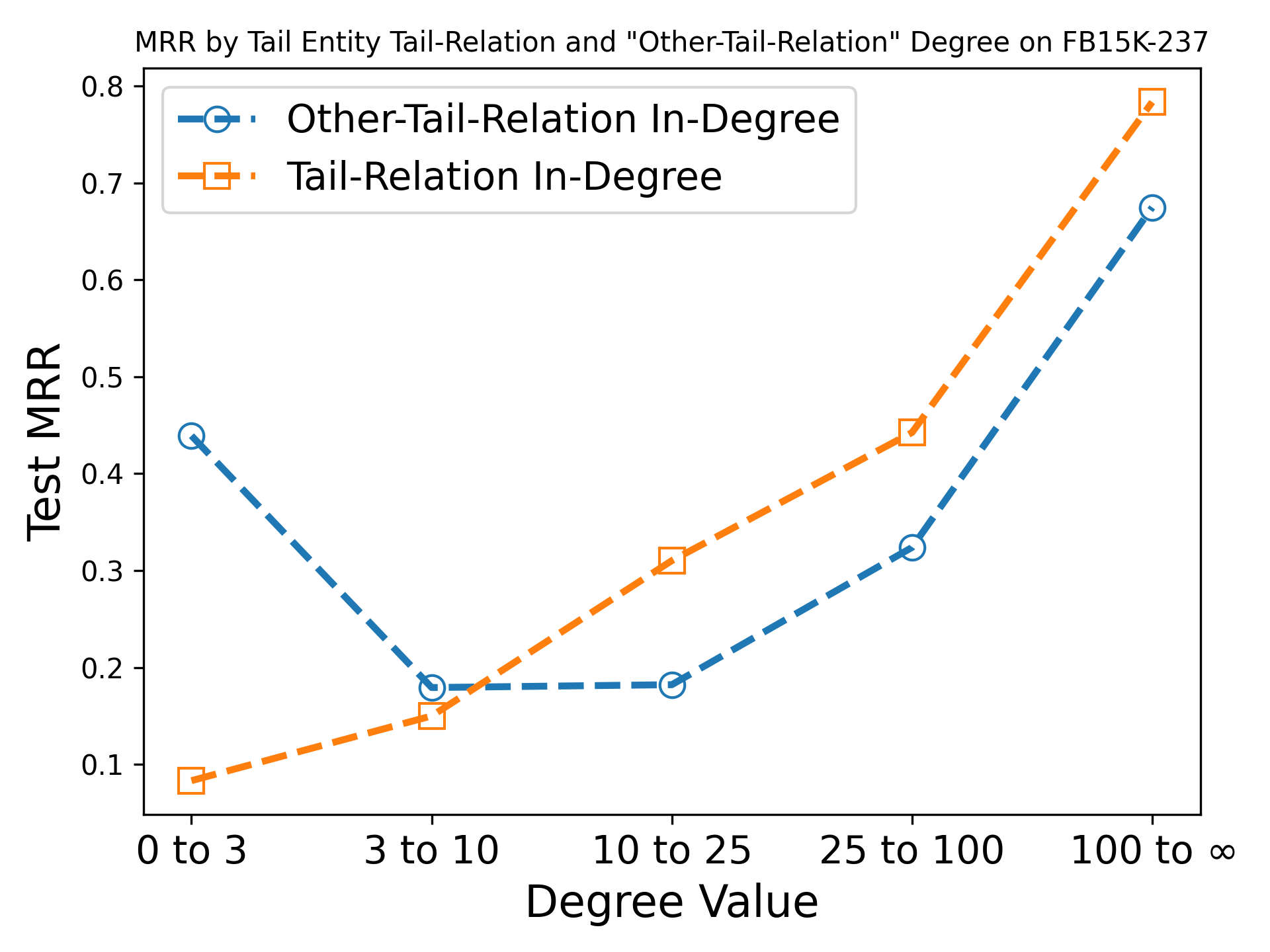}
      \caption{}
      \label{fig:conve_rel_specific_degree}
    \end{subfigure}%
    \begin{subfigure}{.33\textwidth}
      \centering
      \includegraphics[width=1\linewidth]{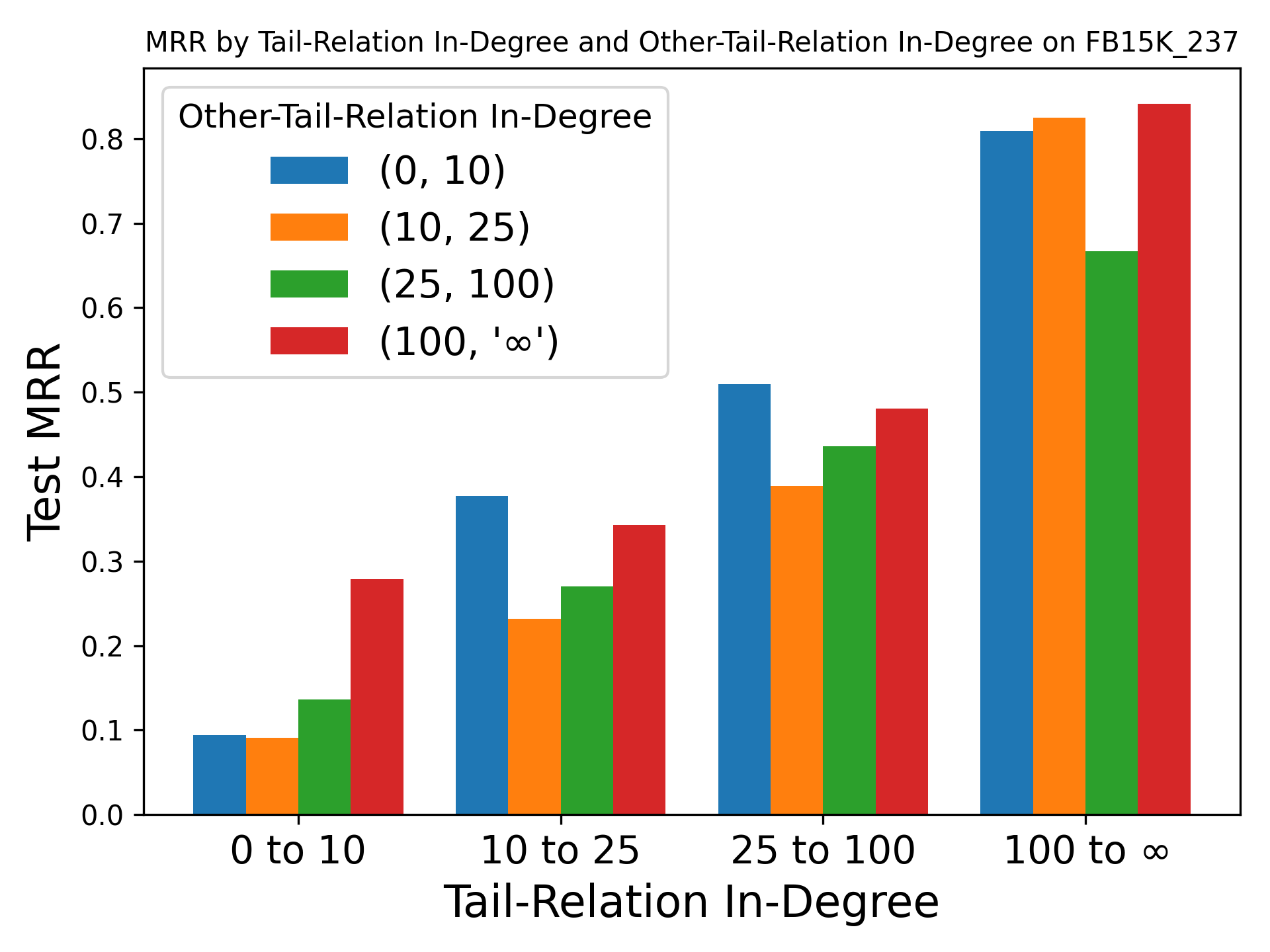}
      \caption{}
      \label{fig:conve_control_degree}
    \end{subfigure}
\caption{MRR when predicting the tail for ConvE on FB15K-237 when we (a)  vary the in-degree and out-degree of the head and tail entity, (b) vary the tail-relation and other-relation in-degree,  and (c) vary the other-relation in-degree for smaller sub-ranges of the tail-relation degree.}
\label{fig:all_analysis_figs_conve}
\end{figure*}

In Section~\ref{sec:prelim} we conduct a preliminary study using TuckER~\cite{tucker} on the FB15K-237 dataset~\cite{fb15k_237}. In this section we further include the corresponding results when using the ConvE~\cite{conve} embedding models. The plots can be found in Figure \ref{fig:all_analysis_figs_conve}. We note that they show a similar pattern to those displayed by TuckER in Figure~\ref{fig:all_analysis_figs}.

\begin{table}[h]
    \small
    \centering
    \caption{Hyperparameter values for ConvE on each dataset.}
    \begin{tabular}{@{}lccc@{}}
        \toprule
            \textbf{Hyperparameter}  & 
            \textbf{FB15K-237} & 
            \textbf{NELL-995 } &
            \textbf{CoDEx-M} \\ 
        \midrule
        Learning Rate   & $1e^{-4}$ & $5e^{-5}$ & $1e^{-5}$ \\
        LR Decay        & None & None  & None  \\
        Label Smoothing & 0 & 0.1 & 0.1 \\
        Dropout \#1 & 0.2 & 0.4 & 0.1 \\
        Dropout \#2 & 0.5 & 0.3 & 0.2 \\
        Dropout \#3 & 0.2 & 0.1 & 0.3 \\
        Degree Threshold & 5 & 5 & 5 \\
        \# Generated & 5 & 5 & 5 \\
        Synth Loss Weight & 1 & 1 & 1 \\
        SWA LR & $5e^{-4}$ & $1e^{-5}$ & $1e^{-5}$\\
        \bottomrule
    \end{tabular}
    \label{tab:conve_hyperparam}
\end{table}

\begin{table}[h]
    \small
    \centering
    \caption{Hyperparameter values for TuckER on each dataset.}
    \begin{tabular}{@{}lccc@{}}
        \toprule
            \textbf{Hyperparameter}  & 
            \textbf{FB15K-237} & 
            \textbf{NELL-995 } &
            \textbf{CoDEx-M} \\ 
        \midrule
        Learning Rate   & $5e^{-5}$ & $5e^{-5}$ & $1e^{-5}$ \\
        LR Decay        & 0.99 & None  & 0.995  \\
        Label Smoothing & 0 & 0 & 0 \\
        Dropout \#1 & 0.3 & 0.3 & 0.3 \\
        Dropout \#2 & 0.4 & 0.3 & 0.5 \\
        Dropout \#3 & 0.5 & 0.2 & 0.5 \\
        Rel Dim & 200 & 100 & 100 \\
        Degree Threshold & 5 & 25 & 5 \\
        \# Generated & 5 & 5 & 5 \\
        Synth Loss Weight & 1 & $1e^{-2}$ & 1 \\
        SWA LR & $5e^{-4}$ & $1e^{-5}$ & $5e^{-4}$\\
        \bottomrule
    \end{tabular}
    \label{tab:tucker_hyperparam}
\end{table}

\subsection{Expected Calibration Error} \label{sec:ece}

Expected calibration error~\cite{guo2017calibration} is a measure of model calibration that utilizes the model accuracy and prediction confidence. Following~\citet{guo2017calibration}, we first split our data into $M$ bins and define the accuracy (acc) and confidence (conf) on one bin $B_m$ as:
\begin{align}
    &\text{acc}(B_m) = \frac{1}{\lvert B_m \rvert} \sum_{i=1}^{\lvert B_m \rvert} \mathbf{1} (\hat{y}_i = y_i), \\
     &\text{conf}(B_m) = \frac{1}{\lvert B_m \rvert} \sum_{i=1}^{\lvert B_m \rvert} \hat{p}_i,  
\end{align}
where $y_i$ is the true label for sample $i$, $\hat{y}_i$ is the predicted label, and $\hat{p}_i$ the prediction probability for sample $i$. A well-calibrated model should have identical accuracy and confidence for each bin. ECE is thus defined by~\citet{guo2017calibration} as:  
\begin{equation} \label{eq:ece}
    \text{ECE} = \sum_{m=1}^M \frac{\lvert B_m \rvert}{n} \lvert \text{acc}(B_m) - \text{conf}(B_m) \rvert, 
\end{equation}
% where $n$ is the total number of samples across all bins. As such, a lower ECE indicates a better calibrated model. For KGC we split the samples into bin by the tail-relation degree. We define the accuracy and confidence for $M$ bins when only predicting the tail as:
% \begin{align}
%     &\text{acc}(B_m) = \frac{1}{\lvert B_m \rvert} \sum_{e=(h, r,t) \in B_m} \text{Hits@10}(e), \\
%      &\text{conf}(B_m) = \frac{1}{\lvert B_m \rvert} \sum_{e=(h, r,t) \in B_m} \sigma(f(h, r, t)),
% \end{align}
% where Hits@10(e) is a boolean variable that is 1 if the tail entity of the triple $e$ is in the top 10 of predicted scores and otherwise 0, $f(h, r, t)$ is a KG embedding score function (e.g. ConvE), and $\sigma$ is the sigmoid function that transforms the output of $f$ into a probability. When calculating the ECE over all samples Eq.~\eqref{eq:ece} is unchanged. When calculating ECE for just one bin of samples (e.g. low degree triples) it is defined as:
where $n$ is the total number of samples across all bins. As such, a lower ECE indicates a better calibrated model. For KGC we split the samples into bin by the tail-relation degree. Furthermore to calculate the accuracy score (acc) we utilize His@10 to denote a correct prediction. For the confidence score (conf) we denote $ \sigma(f(h, r, t))$ as the prediction probability where $f(h, r, t)$ is a KG embedding score function and $\sigma$ is the sigmoid function. When calculating the ECE over all samples Eq.~\eqref{eq:ece} is unchanged. When calculating ECE for just one bin of samples (e.g. low degree triples) it is defined as:
\begin{equation}
    \text{ECE}_m = \lvert \text{acc}(B_m) - \text{conf}(B_m) \rvert,
\end{equation}
where $\text{ECE}_m$ is the expected calibration error for just bin $m$.

\subsection{Proof of Theorem~\ref{th:mixup_reg_form}}
\label{sec:appendix_proof}

In this section we prove Theorem~\ref{th:mixup_reg_form}.
Following recent work \cite{carratino2020mixup, mixup_generalization} we examine the regularizing effect of the mixup parameter $\lambda$. This is achieved by approximating the loss $l_{\theta}$ on a single mixed sample $\tilde{e}$ using the first-order quadratic Taylor approximation at the point $\tau = 1 - \lambda$ near 0. Assuming $l_{\theta}$ is  differentiable, we can approximate $l_{\theta}$, with some abuse of notation as:
\begin{equation} \label{eq:taylor_approx}
    l_{\theta}(\tau) =  l_{\theta}(0) + l_{\theta}'(0) \tau.
\end{equation}
We consider the case where $l_{\theta}$ is the binary cross-entropy loss. Since the label $y_{ij} = 1$ is always true, it can be written as,
\begin{equation}
    l_{\theta}(\tilde{e}) = log \: \sigma \left( f(\tilde{e}) \right),
\end{equation}
where $\sigma$ is the sigmoid function and $\tilde{e} =\left(\tilde{h}, \tilde{r}, t \right)$ is the mixed sample. Since $\tau = 1-\lambda$, we can rewrite the mixed sample as: 
\begin{align}
    \tilde{e}& = ((1 - \tau) h_i + \tau h_j, (1 - \tau) r_i + \tau r_j, t).
\end{align}  
As such, setting $\tau=0$ doesn't mix the two samples resulting in $\tilde{e} = \left(h_i, r_i, t \right)$. The term $l_{\theta}(0)$ is therefore equivalent to the standard loss $ \mathcal{L} (\theta)$ over the samples $S$. We can now compute the first derivative in Eq.~\eqref{eq:taylor_approx}. We evaluate $l_{\theta}'$ via the chain rule,
\begin{equation}
    l_{\theta}' = \frac{\partial \: log \: \sigma \left( f(\tilde{e}) \right)}{\partial \sigma \left(f(\tilde{e}) \right)} \cdot \frac{\partial\sigma \left(f(\tilde{e}) \right)}{\partial f(\tilde{e})} \cdot \frac{\partial f(\tilde{e})}{\partial \tau},
\end{equation}
where $\frac{\partial f(\tilde{e})}{\partial \tau}$ is evaluated via the multivariable chain rule to:
\begin{equation} \label{eq:multivariable_chain}
     \frac{\partial f(\tilde{e})}{\partial \tau} = \left(\frac{\partial f(\tilde{e})}{\partial x_{\tilde{h}}} \frac{\partial x_{\tilde{h}}}{\partial \tau} + \frac{\partial f(\tilde{e})}{\partial x_{\tilde{r}}} \frac{\partial x_{\tilde{r}}}{\partial \tau} + \frac{\partial f(\tilde{e})}{\partial x_t} \frac{\partial x_t}{\partial \tau} \right).
\end{equation}
Evaluating $l_{\theta}'$:
\begin{align}
   l_{\theta}' &= \frac{1}{\sigma \left( f \left(\tilde{e} \right) \right)} \cdot \sigma \left( f \left(\tilde{e} \right) \right) \left( 1 - \sigma \left( f \left(\tilde{e} \right) \right) \right) \cdot \frac{\partial f(\tilde{e})}{\partial \tau}, \\
   &= \left( 1 - \sigma \left( f \left(\tilde{e} \right) \right) \right) \cdot \frac{\partial f(\tilde{e})}{\partial \tau}, \\
   &= \left(1 - \sigma \left( f \left(\tilde{e} \right) \right)\right) \left[
   \frac{\partial f \left(\tilde{e} \right)}{\partial x_{\tilde{h}}} (x_{h_j} - x_{h_i}) + \frac{\partial f \left(\tilde{e} \right)}{\partial x_{\tilde{r}}} (x_{r_j} - x_{r_i}) \right],
\end{align}
where the term related to $x_t$ in Eq.~\eqref{eq:multivariable_chain} cancels out since $\partial x_t / \partial \tau=0$.
Since we are evaluating the expression near $\tau=0$, we only consider the original sample in the score function, reducing the above to the following where $\Delta h = (x_{h_j} - x_{h_i})$, and $\Delta r = (x_{r_j} - x_{r_i})$,
\begin{align}
    l_{\theta}'(0) = \left(1 - \sigma \left( f \left(e_i\right) \right)\right) \left[ \frac{\partial  f \left(e_i\right)}{\partial x_{\tilde{h}}} \Delta h + \frac{\partial  f \left(e_i\right)}{\partial x_{\tilde{r}}} \Delta r\right].
\end{align}
Plugging in $l_{\theta}(0)$ and $l_{\theta}'(0)$ into Eq.~\eqref{eq:taylor_approx} and rearranging the terms we arrive at:
\begin{equation}
    \mathcal{L}_{\text{Mix}}(\theta) =  \mathcal{L} (\theta) + \mathcal{R}_1(\theta)  + \mathcal{R}_2(\theta) ,
\end{equation}
where $\mathcal{R}_1(\theta) $ and $\mathcal{R}_2(\theta) $ are defined over all samples $S$ as:
\begin{align} \label{eq:reg_terms_proof}
   &\mathcal{R}_1(\theta)  =  \frac{\tau}{\lvert S \rvert} \sum_{i=1}^{\lvert S \rvert} \sum_{j=1}^k  \left(1 - \sigma \left( f(e_i) \right)\right)  \frac{\partial f(e_i)^T}{\partial x_{\tilde{h}}} \Delta h, \\
    &\mathcal{R}_2(\theta)  = \frac{\tau}{\lvert S \rvert} \sum_{i=1}^{\lvert S \rvert} \sum_{j=1}^k  \left(1 - \sigma \left( f(e_i) \right)\right) \frac{\partial f(e_i)^T}{\partial x_{\tilde{r}}} \Delta r,
\end{align}
with $\tau = \mathbb{E}_{\lambda \sim \mathcal{D}_{\lambda}} (1-\lambda)$ where $\mathcal{D}_{\lambda} = \text{Beta}(\alpha, \alpha)$.

\end{document}